\documentclass[acmlarge,screen,nonacm]{acmart}
\AtBeginDocument{%
  }

\setcopyright{acmlicensed}
\copyrightyear{2018}
\acmYear{2018}
\acmDOI{XXXXXXX.XXXXXXX}
\acmConference[Conference acronym 'XX]{Make sure to enter the correct
  conference title from your rights confirmation email}{June 03--05,
  2018}{Woodstock, NY}
\acmISBN{978-1-4503-XXXX-X/2018/06}

\newcommand*\captiontype[1]{\def\@captype{#1}}
\newcommand{\urec}{u_{\mathrm{rec}}}

\makeatletter
\def\@ACM@checkaffil{}
\makeatother

\usepackage{framed}

\newenvironment{discussionquote}
  {%
    \MakeFramed{\advance\hsize-\width\FrameRestore}%
    \small
    \setlength{\parindent}{0pt}%
    \setlength{\parskip}{0.5em}%
  }
  {\endMakeFramed}

\usepackage{cleveref}
\usepackage{placeins}
\usepackage{algorithm}
\usepackage{algpseudocode}

\usepackage[most]{tcolorbox}
\newtcolorbox{promptbox}[1][]{
  breakable,
  colframe=black!60,         
  colback=black!5,           
  coltitle=black,            
  title={#1},                
  rounded corners,           
  boxrule=0.5mm,             
  boxsep=5pt,                
  toptitle=1mm,              
  bottomtitle=1mm,           
  left=10pt,                 
  right=10pt,                
  top=5pt,                   
  bottom=5pt,                
  fonttitle=\bfseries        
}

\begin{document}

\title{Learning to Clarify Underspecified Intents Under Limited Interaction}

\author{Pranav M R}
\authornote{Both authors contributed equally to this research.}
\email{pranavmr@mit.edu}
\affiliation{%
  \institution{BITS Pilani \& MIT}
}
\author{Manuel Cherep}
\authornotemark[1]
\email{mcherep@mit.edu}
\affiliation{%
  \institution{MIT}
}

\author{Pattie Maes}
\email{pattie@media.mit.edu}
\affiliation{%
  \institution{MIT}
}

\author{Nikhil Singh}
\email{nikhil.u.singh@dartmouth.edu}
\affiliation{%
  \institution{Dartmouth College}
}

\renewcommand{\shortauthors}{M R, Cherep, Maes, and Singh}

\begin{abstract}
AI assistants receive requests that leave out information needed for a good outcome, for example about users' preferences or goals. They must then either speculate or ask for more information before proceeding. We reconceptualize this as a value-of-information problem: the assistant should acquire information whose absence causes the greatest avoidable loss in user utility. This is rarely known \textit{ex ante}; rather, assistants must \textit{predict} it in order to optimally allocate limited user interactions. We instantiate this problem in image generation and derive a reinforcement learning framework\footnote{\href{https://github.com/PapayaResearch/clarifying}{https://github.com/PapayaResearch/clarifying}} using multi-turn simulated users to maximize utility recovery under uncertainty. In a preregistered study with 456 interactive sessions across 76 human participants, this helped users significantly better match reference images with significantly fewer questions, less total interaction time, and lower cost. This points toward a simple and scalable framework for training language model assistants to better disambiguate user intent by asking more informative questions.
\end{abstract}

\ccsdesc[500]{Computing methodologies~Machine learning}
\ccsdesc[500]{Human-centered computing~Human computer interaction (HCI)}

\keywords{AI Assistants, Agents, Human-AI Interaction, Machine Learning}
\begin{teaserfigure}
 \includegraphics[width=\textwidth]{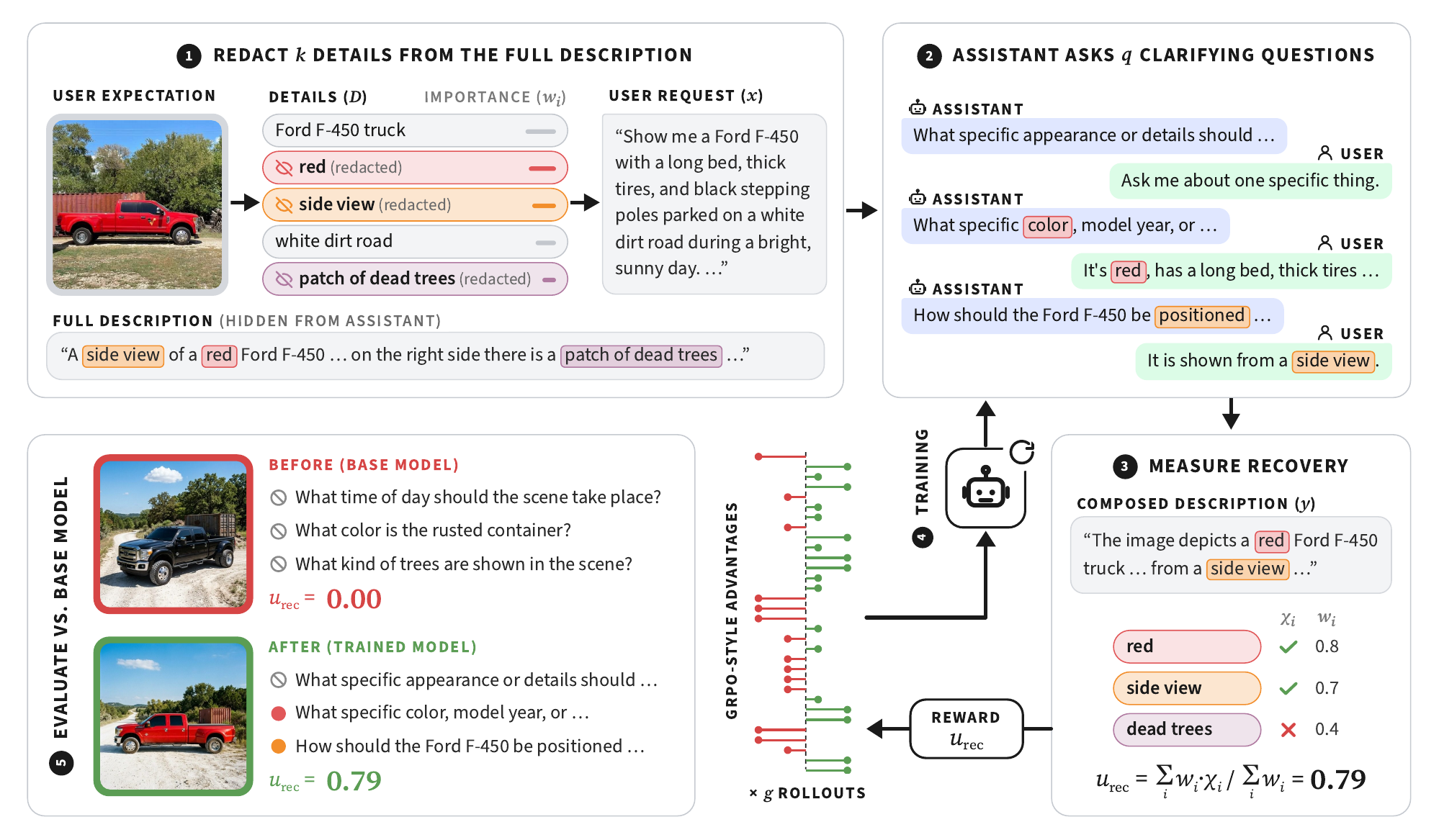}
  \caption{A simplified overview of our training pipeline on a single instance. A set $\mathcal{H}$ of $k$ details is redacted from the full set of details $\mathcal{D}$, and the kept details $\mathcal{D} \setminus \mathcal{H}$ are rewritten into a natural user request $x$. The assistant asks the simulated user clarifying questions and composes, from $x$ and the answers, the description $y$ used to generate the image. The judge then decides which details in $\mathcal{H}$ the assistant recovered by checking $y$ against their checks $\chi_i$, giving the recovered utility $\urec$. Training uses $\urec$ as reward over $g$ rollouts per instance, and the trained assistant is finally evaluated against the base model.}
  \Description{Overview of the training pipeline. A complete reference specification is decomposed into a set of details. A subset of k details is hidden, and the remaining details are rewritten as the initial request x. The assistant receives that request and asks clarifying questions. A simulated user retains the full reference specification and supplies answers. The assistant combines the request and answers into a final description y, which is used for image generation. A judge checks which hidden details are present in y and computes their recovery score. Recovery scores from g rollouts of the same instance supply the training reward. The trained assistant is then compared with the base model. The information available to the simulated user and judge includes the reference details withheld from the assistant.}
  \label{fig:teaser}
\end{teaserfigure}

\maketitle

\section{Introduction}
The cartographers of Jorge Luis Borges' \textit{Del rigor en la ciencia}~\cite{borges1960rigor} are faced with a puzzling trade-off. To make their map perfectly accurate, they must ultimately grow it to the scale of the empire itself. At this point, however, it ceases to be a map in any useful sense. A version of this fate befalls those who seek to delegate to AI systems. Take image generation, for example. Language is inherently imprecise~\cite{sapir1921language,russell1923vagueness}; to truly specify every detail of a desired image (every object, relation, geometry, light and shadow, texture), one might ultimately need to grow the prompt to the scale of the pixels themselves. This is, however, an absurd exercise; such prompts are nearly never needed. Much like the useful maps of \textit{rigor}, good prompts are \textit{abstractive}. They articulate what is essential, and leave implicit what is not. Where we err, then, is not necessarily in saying too little, but also in leaving the wrong things unsaid. To get the best from us, in turn, it would help if our AI assistants~\cite{maes1993learning,kozierok1993learning} could simply ask all the \textit{right} questions.

What makes a given question a good one to ask? Suppose, for instance, that a user requests an image of a family having breakfast. The assistant is faced with virtually infinite possible questions, like: What are they eating? What kind of wood is the table made of? Both of these questions might help to resolve something unclear from the original request. Their \textit{value}, however, may not be equal: this depends on what the user wants the image to accomplish. For example, if it is intended as an ad for a furniture catalog, characteristics of the table are likely more important than those of the food. Selecting from even these two options, let alone the full vast space of possible questions, requires a judgment call about which would be more valuable; a judgment we might like our agents to make well. One way to accomplish this is \textit{clarification}: querying the user about unarticulated aspects of their desired result so that these may be fulfilled.

We study clarification as a value-of-information problem, drawing on decision-theoretic accounts of mixed-initiative interaction and preference elicitation~\cite[cf.][]{horvitz1999principles,boutilier2002pomdp}. Formally, let $U_u(y)$ be the utility of an outcome $y$ to user $u$. Given a conversation history $h$, the value of asking a candidate question $a$ and then acting is:
\begin{equation}
\label{eq:value_of_information}
	V(a \mid h) = \mathbb{E}_{o \mid h,a}\!\left[\max_y \mathbb{E}[U_u(y)\mid h,a,o]\right] - \max_y \mathbb{E}[U_u(y)\mid h]
\end{equation}

The term on the left is the best expected outcome after elicitation, and on the right is the best expected outcome with information already available. The difference is thus the expected benefit arising from clarification.

We construct a framework in which this missing information is known to the training environment but hidden from the assistant. Beginning with a fully specified intent, we decompose it into atomic details, redact a subset of them, and rewrite the remainder as a naturalistic user request. We produce a privileged simulated user which retains access to the complete intent in order to answer assistants' questions. The assistant must then use its limited questions to recover information that was withheld and finally compose its best estimate of the intended result. Because the environment preserves redacted content, we can measure the share of omitted information recovered through the interaction and use it directly as a reward signal.

We instantiate this framework in the special case of text-to-image generation, a suitable first testbed because (1) requests are routinely underspecified, (2) complete specifications can be found in the form of hyper-detailed image descriptions~\cite{onoe2024docci}, and (3) the output is inspectable: we can render the interaction-informed prompt and evaluate the result. We then post-train Qwen3-8B~\cite{yang2025qwen3} against simulated users of varying patience and evaluate the resulting assistants under varying question budgets.

We show, in both simulated and real user experiments, that this training substantially improves output fidelity, that its improvement comes from better interaction rather than better guessing, and that it can under some metrics even improve the efficiency with which results are attained. This pipeline requires no human interaction data beyond the original task descriptions, making it a practical route for building competent assistants that ask before they act.

In all, this work makes the following contributions:
\begin{enumerate}
    \item We develop a value-of-information \textbf{formulation} of clarification for AI assistance, in which questions are valued by their expected effect on the user's eventual outcome under limited interaction.
    \item We introduce a general \textbf{method} for learning natural-language clarification policies from underspecified intent, which includes:
        \begin{enumerate}
            \item A \textbf{dataset-construction procedure} that begins from a complete specification, decomposes it into atomic details, selectively redacts them, and rewrites the remainder as a naturalistic underspecified request.
            \item A \textbf{privileged user-simulation framework} which retains access to the complete intent but selectively reveals information in response to the assistant's questions. This includes simulating varying levels of user patience, which affect the interaction budget under which the assistant must draw its inferences.
            \item A \textbf{reinforcement-learning post-training recipe} that rewards clarification policies according to the utility they recover over multi-turn interactions, using group-relative policy-gradient updates~\cite{shao2024deepseekmath} to learn which questions are most useful to ask.
        \end{enumerate}
    \item We provide \textbf{empirical evidence} that language models can learn substantially better information-acquisition policies through simulated interaction, and that this \textbf{transfers effectively to human user behavior}.
\end{enumerate}

\section{Related Work}
\label{sec:related_work}

\subsection{The value of asking}
\label{sec:value}
Decisions about when and what to ask users have a storied history in intelligent and mixed-initiative systems~\cite{litman1984plan,walker1990mixed,heeman1995collaborating,horvitz1999principles,amershi2019guidelines,zou2023users,clark1986referring}. For instance, Horvitz's second and fifth principles~\cite{horvitz1999principles} discuss ``uncertainty about a user’s goals'' and using, in a cost-aware way, ``dialog to resolve key uncertainties.'' One way to formalize these notions is in terms of preference or utility \textit{elicitation}~\cite{chajewska2000making,boutilier2002pomdp,gajos2005preference}, for example modeling uncertainty over a user's utility as a sequential decision process where queries have costs but answers to them can improve downstream decisions. This accounts for how questions may also be useful because their answers enable useful subsequent questions even if not initially high-value~\cite{boutilier2002pomdp,martin2023model}; this sequential interactive structure is thus essential to our task here, and also demonstrates the complexity of the problem.

The value of asking a question can thus be analyzed via the improvement its answer facilitates~\cite{nelson2005finding,meder2012information,viappiani2010optimal}. This gives a \textit{value-of-information} account of clarification, which is our starting point in this work. This basic idea has been used to query users, for instance ranking candidate questions for StackExchange posts~\cite{rao2018learning} according to how informative their anticipated answers might be, generating questions using answer-conditioned utility models~\cite{rao2019answer}, and evaluating clarification through downstream retrieval performance~\cite{aliannejadi2019asking}. More recently, Zhang et al.~\cite{zhang2025modeling} evaluate such downstream consequences by training language models on preferences informed by simulated future conversation turns. They simulate users with different interpretations of a target ambiguous question, and then prefer the single clarification question that, when provided with an answer from a given simulator, most helps recover the respective interpretation. Recent work has also used language-model-driven interaction to elicit preferences and missing task specifications~\cite{li2025eliciting,qian2024tell,kobalczyk2025active}.

Language models have recently been trained to elicit information through simulated interaction, for instance selecting multi-turn conversations according to the likelihood of an oracle-generated response~\cite{andukuri2024star}, or learning from demonstrations~\cite{qian2024tell}. Broader approaches to collaborative assistance have also been proposed, including optimizing simulated future conversations using rewards that combine task performance, interaction efficiency, and engagement, including in open-ended writing tasks with real users~\cite{wu2025collabllm}. These works help establish that questioning can be fruitfully learned from its downstream consequences. Our focus is then on quantifying the omitted intent which learning can help recover in interactions, and how this recovery depends on the extent of underspecification and level of interaction users are willing to provide.

We build on this line of work by developing a construction that leverages omitted information explicitly in the training and evaluation processes while keeping it hidden from the assistant. This in turn supports (1) measuring utility through satisfaction of intended properties in open-ended generation (where multiple outputs can fulfil the same specification); (2) training multi-turn elicitation policies through reinforcement learning; (3) studying variable interaction budgets, including training with simulated users whose willingness to answer is limited and unknown to the assistant; (4) controlling the degree of underspecification through specification redaction; and (5) evaluating transfer to real users who write their own requests and decide when to stop, measuring result fidelity, interaction burden, and other attributes. Together, these experiments allow us to examine training such assistants within the effort users choose to expend, thus offering a concrete way to incorporate value-of-information objectives into the modern AI assistant post-training stack.

\subsection{Proactive clarification}
\label{sec:proactive}
Perhaps the closest interactive image generation setting to our work comes from Hahn et al.~\cite{hahn2024proactive}. They introduce agents that ask questions before generating an image, using editable belief graphs and prompting principles (e.g. avoiding redundancy). Their simulated users have access to the intended image and can answer questions about it, which allows clarification to be evaluated through, for instance, the generated outputs. Here, questioning behavior is specified through heuristics and prompts. The same applies to CLAM~\cite{kuhn2022clam} outside image generation, as it classifies whether a user's request is ambiguous and only then prompts the model to ask a clarifying question. The decision of whether and what to ask is based on few-shot prompting rather than optimization. Recently, A-MPQC~\cite{wang2026agentic} studies a similar problem in design generation, and accounts for how questions might differ in the effort required to answer them. Here, the agent adapts its questioning strategy as the user answers or rejects questions, in part using a curriculum of question types including free-response, multiple-choice, and binary formats that require different levels of user effort.

We build on the simulated-user approach implicit in these works, along with the value-of-information principle discussed previously, to explicitly train AI assistants to perform better at this type of clarification question generation by deriving a measurable training objective. Our setup also facilitates more systematic comparison of different question budgets, levels of underspecification, and potential user behaviors.

\subsection{Preference feedback}
\label{sec:preference}
Effective delegation also requires decisions about how much of the work of specifying outcomes resides with the user. In image generation, PromptCharm~\cite{wang2024promptcharm} supports users in refining their own prompts, but does not model what the user wants, so the burden of recognizing what is missing remains with the user. APPO~\cite{li2026preference} redistributes this work partially to the system through an adaptive, preference-informed prompt optimization method in which users select preferred images and the optimizer infers useful prompt changes from those selections. APPO assumes, implicitly, that the user's initial prompt contains essential elements of the desired image, thus allowing subsequent search to focus largely on implicit preferences. They state that ``recovering missing key elements (e.g., the primary object in an image) remains challenging due to the vast search space.'' 

Our work also assigns to the assistant the responsibility for identifying and acquiring such properties; we explicitly allow important details to be absent from the initial request, and train the assistant to elicit information that improves recovery of such details. We believe this is thus complementary to APPO's interactive optimization and could be meaningfully combined to scaffold multiple levels of outcome specification, however we note that this is speculative and leave investigating such potential combinations to future work.

\subsection{Learning via simulated interaction}
\label{sec:learning}
Finally, our work deals with teaching an AI assistant to interact more effectively with a user. Learning useful interactive behavior normally requires repeated opportunities to act, observe results, and improve on the basis of those observations. A natural formulation of this is as a (partially observable) Markov Decision Process (MDP) (as discussed in \Cref{sec:value}~\cite{boutilier2002pomdp,singh2002optimizing}). \textit{Reinforcement learning} offers algorithms for learning to improve such sequential decisions through their eventual rewards. The issue here is that learning in this fashion can require many trial interactions (i.e. it is sample-inefficient), each demanding user participation and, for language-model assistants, this is compounded by the substantial computation involved in generating dialogue and updating many parameters. User simulation makes such experimentation feasible at scale without requiring people to participate in every training interaction~\cite{shi2019build,sekulic2022evaluating}, a principle that dates back to early work on learning dialogue policies~\cite{levin1997stochastic,schatzmann2007agenda}.

In HCI and specifically \textit{computational interaction}, reinforcement learning has been used both to construct models of user behavior~\cite{hetzel2021complex} and to optimize interface adaptations using predicted or simulated user responses~\cite{todi2021adapting,langerak2022marlui,chen2015emergence,o2015designscape}. These approaches thus allow interaction strategies to be explored before evaluating their performance with people, an opportunity which has recently been extended to language-model agents, for example learning improved social interaction~\cite{wang2024sotopia}.

Most relevant for us, UserRL~\cite{qian2025userrl} involves training assistants using multi-turn interactions with simulated users, studying how different task success rewards improve assistance in settings like travel planning and puzzle solving. We extend this line of work to open-ended generation, where the output space is far more complex and consequential details may be missing from the initial request. We deliberately withhold aspects of a complete specification to efficiently train multi-turn clarification under systematically varying underspecification and interaction budgets, and evaluate the resulting assistants with users, measuring both output quality and interaction burden. We also emphasize the assistant's impact on real users, while \cite{qian2025userrl} reports a preliminary study with only five participants and a simplified protocol that is vulnerable to data leakage.

\section{Methods}
\label{sec:methods}

We define a general framework in which a user's utility is known but hidden from an AI assistant, train the model underlying the assistant by rewarding it with the recovered utility $\urec$ obtained through interacting with simulated users (who may or may not run out of patience), and evaluate the trained assistant with real users in a study.

We begin with a fully specified intent (here, a highly detailed image caption) and redact $k$ details to generate a partial intent in which missing information (i.e., missing utility) is known exactly. Thus, what the assistant recovers can be compared against ground truth. Besides the assistant $\pi_\theta$, we use four language model instances in our pipeline, each prompted to accomplish a different piece of the task: a decomposition model and a redaction model that we use to build task instances (full specifications are in \Cref{app:synthesis}), a user simulator model $\pi_u$, and a judge $\mathcal{J}$ which is used to compare recovered prompts against ground truth. \Cref{sec:setup} describes the setup first in the general case, and then for the image generation task we study. All prompts are listed in \Cref{app:prompts}.
  
\subsection{Problem setup}
\label{sec:setup}

\paragraph{Intent and utility.} Let $s^\star$ denote a complete natural language specification of the task the user wants done. The decomposition model transforms it into $m$ atomic details $\mathcal{D} = \{d_1, \dots, d_m\}$, where each $d_i = (\delta_i, w_i, \chi_i)$ includes the content $\delta_i$ itself, an associated importance weight $w_i \in [0, 1]$ (how much the detail is expected to influence the outcome of the task), and a check $\chi_i : \mathcal{Y} \to \{0, 1\}$, which is a binary question about $d_i$ that the judge uses as an evaluation signal (e.g. ``Is the car red?'' or ``Is there a cat on the left side?''). The assistant's outcome is $y \in \mathcal{Y}$, the specification it composes after clarifying with the user, from which the task's output is produced. In this framework, we model the user's utility from $y$ as the importance-weighted share of the full task specification (here, represented as the set of details) it contains:
\begin{equation}
\label{eq:utility}
U_u(y) = \frac{\sum_{i=1}^{m} w_i\, \chi_i(y)}{\sum_{i=1}^{m} w_i}.
\end{equation}

\paragraph{Redaction.} For each task instance, we hide some portion of the $m$ details from $s^\star$, a set $\mathcal{H} \subset \mathcal{D}$ with $|\mathcal{H}| = k$ drawn sequentially without replacement with probability proportional to the importance weight $w_i$ (see \Cref{app:synthesis}). The count $k$ is itself drawn per instance uniformly from $\{1, \dots, K\}$ for a set-wide maximum $K$, so that instances differ in the level of underspecification to avoid the assistant relying on a fixed number of queries. The redaction model then writes the request $x$, a concise natural request that contains only those retained details $\mathcal{D} \setminus \mathcal{H}$. The assistant observes only $x$ and is tasked with querying $\pi_u$ to recover as much of $\mathcal{H}$ as possible.

\paragraph{Dialogue.} The assistant $\pi_\theta$ is a language model instance instrumented with three task-specific prompts leading to three instantiations which toggle depending on the state of the interaction: $\pi_\theta^{\mathrm{ask}}$, $\pi_\theta^{\mathrm{decide}}$ and $\pi_\theta^{\mathrm{compose}}$. The three share the same weights and receive the same dialogue history so far, and differ only in their system prompt and in what they emit: $\pi_\theta^{\mathrm{ask}}$ asks the user a clarification question, $\pi_\theta^{\mathrm{decide}}$ decides whether to ask another, and $\pi_\theta^{\mathrm{compose}}$ writes the final outcome $y$ once questioning ends. $\pi_u$ denotes a privileged user simulator, which has access to $s^\star$ but is instructed to carefully answer only what is actually asked; i.e., it is guided to never volunteer unasked details and to avoid giving responses that are not clearly supported by $s^\star$. The assistant asks atomic queries, questions that $\pi_u$ should, in the ideal case, answer with at most one detail.

An episode begins with history $h_0 = x$. At turn $t$ the assistant asks $a_t \sim \pi_\theta^{\mathrm{ask}}(\cdot \mid h_{t-1})$, the user answers $o_t = \pi_u(a_t, h_{t-1}; s^\star)$, and the history is updated as $h_t = h_{t-1} \oplus (a_t, o_t)$. We denote the number of questions answered in an episode by $q$, with the episode ending when either the assistant decides not to ask another question or the user declines to answer additional questions. The assistant then composes $y \sim \pi_\theta^{\mathrm{compose}}(\cdot \mid h_q)$, the specification of what it infers the user wants. This is the assistant's best response to its own implicit estimate $\widehat{U}_u(\cdot \mid h_q)$ of the user's utility, formed from $x$ and the answers $\{o_1,\ldots, o_t\}$, and then $U_u(y)$ can be used to score it against the ground truth.

\paragraph{Recovered utility.} We split the sum in \Cref{eq:utility} into retained vs. redacted details. The retained details are in $x$ and the assistant is instructed to preserve them, so we treat their contribution to $U_u(y)$ as constant. What varies then is the redacted part, and normalizing it gives the recovered utility:
\begin{equation}
\label{eq:urec}
\urec(y) = \frac{\sum_{i \in \mathcal{H}} w_i\, \chi_i(y)}{\sum_{i \in \mathcal{H}} w_i} \in [0, 1],
\end{equation}
the share of the missing utility (which, in our model, the user desires but did not explicitly state) that the assistant recovers after interaction. Under the assumption above that the retained details are preserved, \Cref{eq:utility} is an increasing affine function of this share, $U_u(y) = c + (1 - c)\cdot\urec(y)$ with $c = \sum_{i \notin \mathcal{H}} w_i / \sum_{i=1}^{m} w_i$ the retained share of the weight, so maximizing $\urec$ on an instance maximizes the user's utility. We compute it under both importance weights $w_i$ and uniform weights ($w_i = 1 \;\forall i$), and report the uniform-weighted value throughout as a simple and interpretable baseline (importantly, they correlate at 0.985 or above and rank every evaluation identically, more details in \Cref{app:training}).

\paragraph{Asking modes.}
Within any interaction, an assistant can adopt a number of different policies related to how much to ask. We simplify this to three distinct policy types, or \textit{asking modes}:
\begin{itemize}
  \item \textbf{No-asking}: the assistant asks no questions, composing $y$ from $x$ + any implicit-detail inference alone. This measures what the request and the assistant's prior already yield, the floor of $\urec$.
  \item \textbf{Fixed-asking}: the assistant asks exactly $k$ questions (one per hidden detail) so an optimal policy can recover every hidden detail. Here, it chooses what to ask but not how much.
  \item \textbf{Open-asking}: before each question the assistant decides via $\pi_\theta^{\mathrm{decide}}$ whether to ask another one. Questioning terminates at the first stop decision or upon exhausting the budget, whichever occurs first. We set the budget to $2k$, so the assistant can both under-ask and over-ask.
\end{itemize}

The assistant knows neither the hidden count $k$ nor the budget of its mode in advance.

\paragraph{Users.}
In addition to different asking behaviors, we consider variation in answering behavior. Intuitively, just because an assistant asks a question does not mean the user is obliged to answer it. They may abandon the interaction at any time, or abandon by way of refusing to answer additional questions before production. We thus design two forms of user simulator which differ only in when they stop answering the assistant's questions. The \textbf{patient} user answers every question, and the \textbf{impatient} user can stop answering at any point. To simulate this latter behavior, we draw its patience $p$, the number of questions it answers before asking for the outcome, from:
\begin{equation}
\label{eq:patience}
\Pr(p \le b) = \left(\frac{b}{B}\right)^{\alpha}, \qquad b \in \{1, \dots, B\}, \quad \alpha > 0.
\end{equation}
Here $B$ is the budget of the asking mode, $B = k$ under fixed asking and $B = 2k$ under open asking. Since a uniform stopping distribution corresponds to $\alpha = 1$, we choose $\alpha > 1$ so that the user's patience follows a waning schedule; i.e. the chance of a question being the last one answered grows as the conversation carries on, as we might intuitively expect of real users. In our experiments, we heuristically set $\alpha = 1.5$, an intermediate value, so the user answers roughly two thirds of the questions on average if the assistant uses its whole budget.

\paragraph{Combinations of asking modes and user simulator behaviors.}
The 3 asking modes $\times$ 2 user simulators give the following conditions, with the answered count $q$ of each being:
\begin{center}
\begin{tabular}{lll}
  \toprule
  & Patient user & Impatient user \\
  \midrule
  No-asking & $q = 0$ & $q = 0$ \\
  Fixed-asking & $q = k$ & $q = p$ \\
  Open-asking & $0 \le q \le 2k$ & $0 \le q \le p$ \\
  \bottomrule
\end{tabular}
\end{center}

\paragraph{Image generation.} We instantiate the setup for image generation (\Cref{fig:teaser}) using the DOCCI~\cite{onoe2024docci} dataset. We choose DOCCI because of its high-density, human-annotated descriptions that target key challenges in image generation such as spatial reasoning or world knowledge. The specification $s^\star$ is one of the hyper-detailed image descriptions in the dataset, each written for a photograph that we keep as the reference image. The outcome $y$ is the image description composed by the assistant, which a text-to-image model $\mathcal{G}$ renders into the image $\mathcal{G}(y)$ the user receives. Since we focus on elicitation, we first score $y$ directly and separately verify that improvements in $y$ carry over to the rendered images $\mathcal{G}(y)$. We set $K = 15$ as the maximum number of redactions, take a subset of 2,000 DOCCI descriptions, and keep those that decompose into at least $K + 5$ details (so that there is enough residual structure to produce reasonable starting prompts). This produces 479 instances, which we split by source into 290 for training, 111 for validation, and 78 for testing.

\subsection{Training to maximize the recovered utility}
\label{sec:training}

\paragraph{Models.} The model we use as the assistant is Qwen3-8B~\cite{yang2025qwen3} with reasoning on; we use it because it is a popular choice for fine-tuning\footnote{Qwen3-8B is both a highly popular choice for fine-tuning (\href{https://huggingface.co/blog/state-of-open-models-summer-2026}{huggingface.co/blog/state-of-open-models-summer-2026}) and a highly effective one (\href{https://www.distillabs.ai/blog/what-small-language-model-is-best-for-fine-tuning/}{distillabs.ai/blog/what-small-language-model-is-best-for-fine-tuning}).}. For every question, decision and description, the model has a sampling cap of 4,096 tokens shared by the reasoning and visible text, and only the visible text becomes part of the history and is given to the user and the judge. The remaining models use Gemini 3.5 Flash-Lite, a popular and inexpensive workhorse model. As a supplementary analysis, we audit the simulator's answers with a judge to check how often they state about one hidden detail each and only when asked (\Cref{app:audit}).

\paragraph{Objective and update.} Let $\tau = (h_q, y)$ denote an episode and $r(\tau) = \urec(y)$ its reward, the uniform-weighted version of \Cref{eq:urec}. We maximize the expected reward $\mathbb{E}_{x}\, \mathbb{E}_{\tau \sim \pi_\theta(\cdot \mid x)}[r(\tau)]$ using policy gradients over rollout groups (\Cref{alg:training}). This objective approximates \Cref{eq:value_of_information} summed over the episode with $\pi_u$ in place of the user and $\urec$ in place of $U_u$ (\Cref{eq:urec}). The per-question values telescope to the expected recovery (with optimal composition) after the dialogue minus that of $x$ alone, a constant.  Assuming the assistant's composition function is near-optimal, maximizing the expected reward therefore approximately maximizes the total value of information. Each iteration draws $M = 8$ training instances and $g = 32$ episodes per instance, judges every episode, and centers the rewards within each instance's group, $A_j = r_j - \frac{1}{g} \sum_{j'=1}^{g} r_{j'}$, a GRPO-style advantage function~\cite{shao2024deepseekmath} without standard-deviation normalization. Centering helps to debias the reward signal, since details guessable from $x$ alone are in expectation recovered by multiple rollouts alike, so the update rewards questions that helped recover more relative to this group average. Under the impatient user, $p$ is drawn once per group, so the rollouts of one instance compete under the same patience budget. We then use an importance-sampled policy-gradient step to update low-rank adapter (LoRA)~\cite{hu2022lora} parameters $\theta$ (an efficient fine-tuning approach). We set LoRA hyperparameters as rank 32 and learning rate $4 \times 10^{-5}$, with $A_j$ applied to every token the assistant sampled in episode $j$, reasoning included.

\begin{algorithm}[t]
\caption{Training an assistant on recovered utility}
\label{alg:training}
\begin{algorithmic}[1]
\Require instances $(x, \mathcal{H}, s^\star)$ built by redaction (\Cref{app:synthesis}), assistant $\pi_\theta$, user simulator $\pi_u$, judge $\mathcal{J}$, iterations $N$, instances per iteration $M$, rollouts per instance $g$
\For{$N$ iterations}
  \State sample $M$ training instances
  \For{each instance $(x, \mathcal{H}, s^\star)$ with $k = |\mathcal{H}|$}
    \State draw the user's patience $p$ once for the group \Comment{$p = k$ if patient, else \Cref{eq:patience} with $B = k$}
    \For{$j = 1, \dots, g$} \Comment{the rollout group of the instance}
      \State $h \gets x$
      \For{$t = 1, \dots, p$} \Comment{fixed-asking, one question per hidden detail}
        \State $a_t \sim \pi_\theta^{\mathrm{ask}}(\cdot \mid h)$, \quad $o_t \gets \pi_u(a_t, h; s^\star)$, \quad $h \gets h \oplus (a_t, o_t)$
      \EndFor
      \State $y_j \sim \pi_\theta^{\mathrm{compose}}(\cdot \mid h)$
      \State $r_j \gets \urec(y_j) = \frac{1}{k} \sum_{i \in \mathcal{H}} \mathcal{J}(\chi_i, y_j)$ \Comment{\Cref{eq:urec} under uniform weights}
    \EndFor
    \State $A_j \gets r_j - \frac{1}{g} \sum_{j'=1}^{g} r_{j'}$ for every $j$ \Comment{group-centered advantage}
  \EndFor
  \State one policy-gradient step on $\theta$, $A_j$ applied to every token the assistant sampled in rollout $j$
\EndFor
\State \Return the checkpoint with the best validation $\urec$ against $\pi_u$
\end{algorithmic}
\end{algorithm}

\paragraph{Training against patient and impatient users.} As noted, we train against two versions of the user simulator. Both are trained in fixed-asking mode with $B = k$ and $r(\tau) = \urec(y)$, so that the gradient pushes the assistant toward better clarifying questions at the same budget. The \textbf{patient-trained} assistant's reward is what it was able to recover with $k$ answered rounds, and the \textbf{impatient-trained} assistant is rewarded in proportion to what it could recover within an unknown time limit, which we hypothesize could learn to favor asking the most valuable questions first. Each run trains up to 300 iterations from the base weights, with early stopping on reasoning length, and the best checkpoint of each run is chosen on validation recovery under the patient or the impatient user respectively (\Cref{app:training}). The untrained weights in the same modes are the \textbf{base} model.

\subsection{User study}
\label{sec:userstudy}

We ran a preregistered study\footnote{Pre-registration on AsPredicted \#310,205 (anonymized link: \href{https://aspredicted.org/mx4wx3.pdf}{aspredicted.org/mx4wx3.pdf})} to validate the effectiveness of the trained assistant with real users who can decide to stop answering questions at any point. The study was determined to be exempt by our institutions' IRBs.

\paragraph{Task.} In each round, the participant sees one target image out of 78 possible test instances. The task consists of writing a prompt in order to maximally reconstruct the seen reference image, and answering the assistant's clarifying questions one at a time, for as long as they like, up to the global cap of $K = 15$ questions. The image is only generated once, at the end of the round, from the description the assistant composes, which the participant does not directly observe. The real user here plays the role of the impatient simulated user during training, but using their authentic human impatience. To incentivize them to iterate (and not always click through to the final result to save time), we offered a bonus based on performance (discussed below).

\paragraph{Design.} The within-subjects study compares two models: the base model and the patient-trained assistant (we discuss the rationale for this in the simulated results section). Both were run with the prompts and settings of \Cref{sec:training} in fixed-asking form, but since we have no explicit redactions here, they are fixed to ask as many of the $K = 15$ questions as the user will allow. Therefore, the assistant asks a question every turn and the participant decides when to stop. Each participant completes six such task rounds (split evenly by model, in randomized order), on six target images drawn at random without replacement.

\paragraph{Recruitment.} Participants were recruited on Prolific; screened for age, English as a native language, and normal or corrected vision. We compensated them at \$12/hour, plus a \$2 bonus based on performance to incentivize them as described above (we ultimately awarded the bonus to all). All participants had to pass a task comprehension check before the first round. Additional details are given in \Cref{app:userstudy}.

\subsection{Evaluation and analysis}
\label{sec:analysis}

We evaluate clarification primarily through detail recovery, changes in similarity to the reference images/captions, and features of interaction required to achieve these outcomes (e.g. number of questions). We first analyze results with simulated users on test-set image instances, which allows us to establish that the training has the intended effect, and to conduct comprehensive controlled comparisons across assistants, asking policies, and user patience levels. Importantly, this data has a known number of redactions, so we can directly estimate the utility gain as defined previously.

We then analyze our human evaluation, as described, to study how the trained assistant performs with real users in interactive tasks. We use a similar estimation and inference approach across these evaluations.

\subsubsection{Evaluation conditions}
The simulated evaluation consists of 78 held-out (test set) DOCCI image sources, with 1--15 redacted details per instance. We compare the Qwen3-8B base model with the patient-trained and impatient-trained assistants as described previously. We test each of these assistants in two regimes: fixed asking $B = k$ (recall that $k$ is the number of redactions), and open asking $B = 2k$, with both patient and impatient simulated test users. We additionally evaluate each assistant in a no-asking mode to establish a baseline (in this condition, the training shouldn't confer a significant advantage). This gives 3 assistants (base/patient-trained/impatient-trained) $\times$ 2 conditions (fixed/open) $\times$ 2 user-simulators (patient/impatient) = 12 asking conditions + 3 no-asking conditions (corresponding to the 3 assistants), for a total of 15 evaluation conditions.

In human interactions, we compare only the base and patient-trained assistants.

\subsubsection{Outcome measures}
In the simulated study, our primary outcome is recovered utility (with equal weights for all details; we additionally test importance-weighted results as a robustness check). We express this as percentage of hidden details recovered in the final description, and report changes in absolute terms (percentage points). Since recovery can include both information acquired through answers as well as simply correct \textit{ex ante} inferences, the no-asking conditions estimate performance without dialog. In the human study, the number of redactions is implicit and never precisely known, and so we do not measure this outcome there.

Across both evaluations, we measure \textbf{semantic change} using unit-normalized embeddings from Gemini Embedding 2 (denoted $e(\cdot)$ below). For an initial output $z_\text{init}$, final output $z_\text{fin}$, and reference $z_\text{ref}$, we compute change in cosine similarity:

\begin{equation}
	\Delta_{\mathrm{sim}}(z_\text{init},z_\text{fin};z_\text{ref}) = \text{sim}(z_\text{fin}, z_\text{ref}) - \text{sim}(z_\text{init}, z_\text{ref}) = e(z_\text{fin})^\top e(z_\text{ref}) - e(z_\text{init})^\top e(z_\text{ref}), \qquad \|e(z)\|_2 = 1
\end{equation}

We apply this across multiple modality conditions. For the prompt text, we compare the redacted request and final description with the complete reference description. For images, we compare images generated from those two texts with the reference photograph (upscaled for machine study, original for human study since it is what they see and try to match). We also compute both cross-modal comparisons: text against reference image, and generated image against reference text.

We also log interaction measures: question-answer counts, question and answer word counts, and words per answered question and answer. The human interactions additionally give us wall-clock durations, and we measure token usage as a proxy for cost. Together, these measures help to distinguish the number of exchanges from the effort and time they require. The per-question and per-answer measures exclude logged interactions without any answered questions.

\subsubsection{Regression analysis}
We then estimate condition differences using OLS regressions with reference-image fixed effects (to account for stable differences between tasks) and cluster-robust standard errors. We cluster standard errors by source image in simulated evaluation, where conditions share the same redacted request, and by participant in human evaluation, where participants compose their own requests but complete multiple tasks. We adjust for answered-question count, except when question count or a count-derived measure is the outcome, and in the simulated-performance overviews (\Cref{fig:recovery,fig:embedding_scores}), which retain source-image fixed effects but omit question-count adjustment (since there is a no-asking condition). In the plot annotations, we use $p$-values derived from contrasts between conditions from these models. Tests and contrast confidence intervals use a $t(C-1)$ reference distribution, where $C$ is number of clusters. We adjust $p$-values using the Benjamini--Hochberg procedure to control the false discovery rate across the specified control comparisons within each simulated evaluation outcome, and separately across quality and process outcomes in the human evaluation.

For condition comparisons, we plot estimated marginal means (EMMs), with all reference images weighted equally and answered-question count at its pooled mean wherever included as a covariate. The error bars show 95\% confidence intervals using the model's cluster-robust covariance matrix and $t(C - 1)$ critical values, and significance stars are BH-adjusted contrasts between these EMMs.

For simulated asking interactions, we fit additive, two-way, and three-way specifications to examine how assistant differences depend on features like asking policy, user patience, and dialogue measures (e.g. question count). For the human quality models, we similarly compare specifications with and without question-count adjustment and also image fixed effects.

\section{Results}
\label{sec:results}

\subsection{Model Results}
Unless otherwise stated, differences discussed below are regression-adjusted contrasts with 95\% confidence intervals and BH-adjusted $p$-values.

\subsubsection{Can models learn to ask more effective questions?}
\begin{figure*}[!htb]
    \centering
    \includegraphics[width=\linewidth]{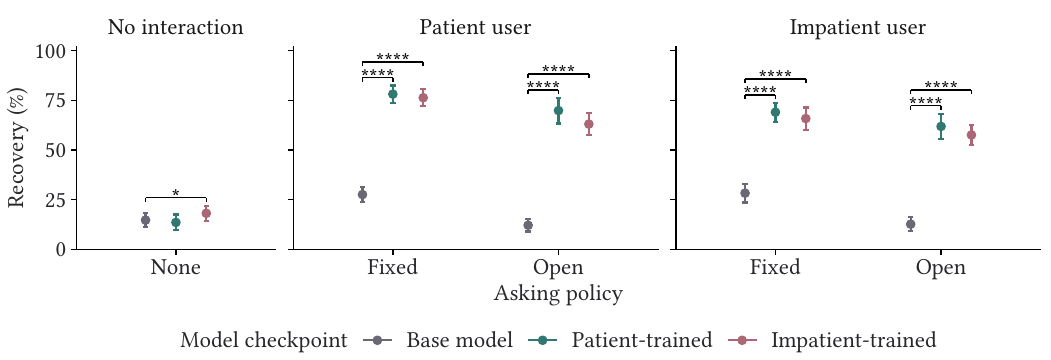}
    \caption{Recovered utility in simulated evaluation, measured as percentage of hidden details recovered by the final description. Points show estimated marginal means; bars are 95\% source-image-clustered regression confidence intervals. Brackets compare trained checkpoints vs. base using source-image fixed effects and source-image-clustered standard errors. Asterisks denote BH-adjusted significance: $^{*}p<.05$, $^{****}p<.0001$.}
    \Description{Three panels comparing no-interaction, patient simulated users, and impatient simulated users. Recovery is low for all checkpoints without interaction with p<0.05 for base vs. impatient-trained. Both trained checkpoints have substantially higher recovery than base under fixed and open asking for both patient and impatient simulated users, all p<0.0001.}
    \label{fig:recovery}
\end{figure*}

We find that training increases recovered utility in all asking conditions (\Cref{fig:recovery}). Across all eight trained-versus-base comparisons, adjusted gains range from image-adjusted gains range from 37.6 to 57.7 pp., all $p_{\mathrm{BH}}<.0001$. With a patient user and fixed asking, the patient-trained checkpoint gains 50.6 points [43.8, 57.3] and the impatient-trained checkpoint gains 48.7 points [42.4, 55.0]. Observed recovery is 27.7\% for base, 78.2\% for patient-trained, and 76.4\% for impatient-trained.

Gains persist with impatient simulated users. Under fixed asking, adjusted gains are 40.7 points [33.3, 48.2] for patient-trained and 37.6 points [30.0, 45.2] for impatient-trained. Under open asking, the corresponding gains are 57.7 and 50.9 points with patient users, and 49.2 and 44.9 points with impatient users. The gains are also robust across alternative specifications (\Cref{tab:recovery_regressions}).

Without asking, these effects largely disappear. Patient-trained differs from base by $-1.2$ points [$-3.5$, 1.2], $p_{\mathrm{BH}}=.354$; impatient-trained gains 3.4 points [0.3, 6.5], $p_{\mathrm{BH}}=.041$. Observed recovery is 14.8\%, 13.6\%, and 18.2\% for base, patient-trained, and impatient-trained, respectively. The gains from asking thus support the interpretation that the benefit arises from learned interaction, not just e.g. improved zero-shot inference of missing details.

\subsubsection{Do better questions lead to better output fidelity?}
\begin{figure*}[!htb]
    \centering
    \includegraphics[width=\linewidth]{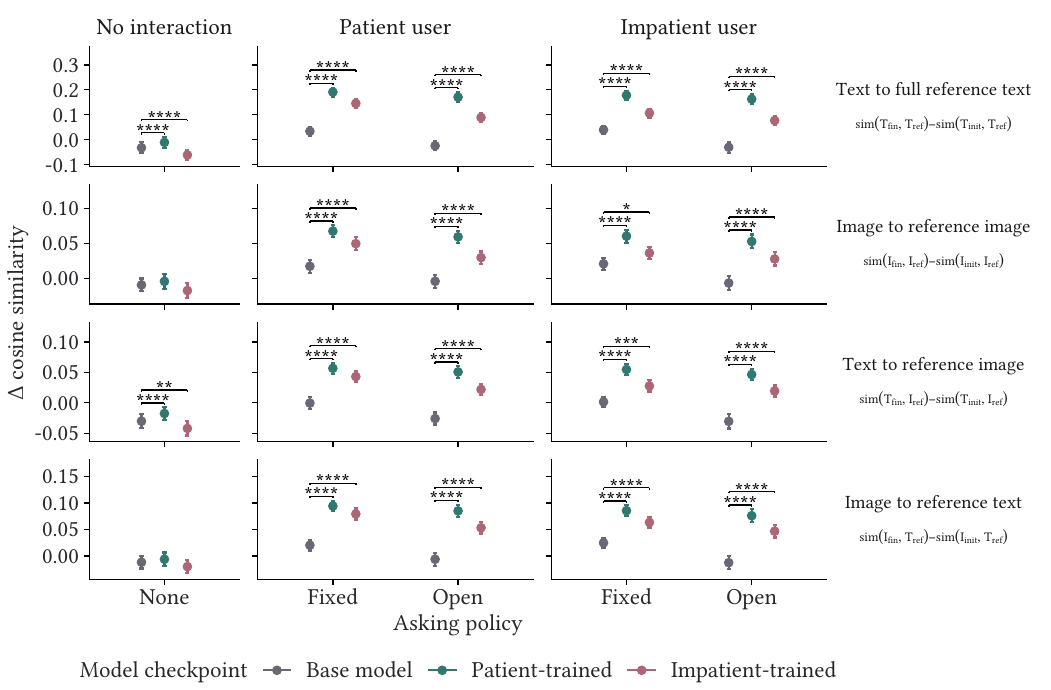}
    \caption{Change in cosine similarity across text/image comparisons in simulated evaluation, using complete descriptions and upscaled source images as references. Positive values indicate improvement over the initial request or its generated image. Points show estimated marginal means; bars are 95\% source-image-clustered regression confidence intervals. Brackets show trained-vs.-base contrasts with source-image fixed effects and source-image-clustered uncertainty. BH-adjusted significance: $^{*}p<.05$, $^{**}p<.01$, $^{***}p<.001$, $^{****}p<.0001$.}
    \Description{4 rows showing text-to-text, image-to-image, text-to-image, and image-to-text similarity gains across no-interaction and two simulated-user types. Trained checkpoints generally have larger gains under both asking policies. Patient-trained has the highest adjusted means across the asking comparisons.}
    \label{fig:embedding_scores}
\end{figure*}

Training also improves reference-similarity gains across all four text/image comparisons (\Cref{fig:embedding_scores}). The image-adjusted advantages are positive in every asking condition, ranging from 0.040--0.195 for patient-trained and 0.016--0.113 for impatient-trained. All comparisons are significant at $p_{\mathrm{BH}}<.05$. The alternative specifications largely reproduce this pattern (\Cref{tab:semantic_regressions,tab:crossmodal_regressions}).

\subsubsection{Can models decide how much information to elicit?}
\begin{figure*}[!htb]
    \centering
    \includegraphics[width=\linewidth]{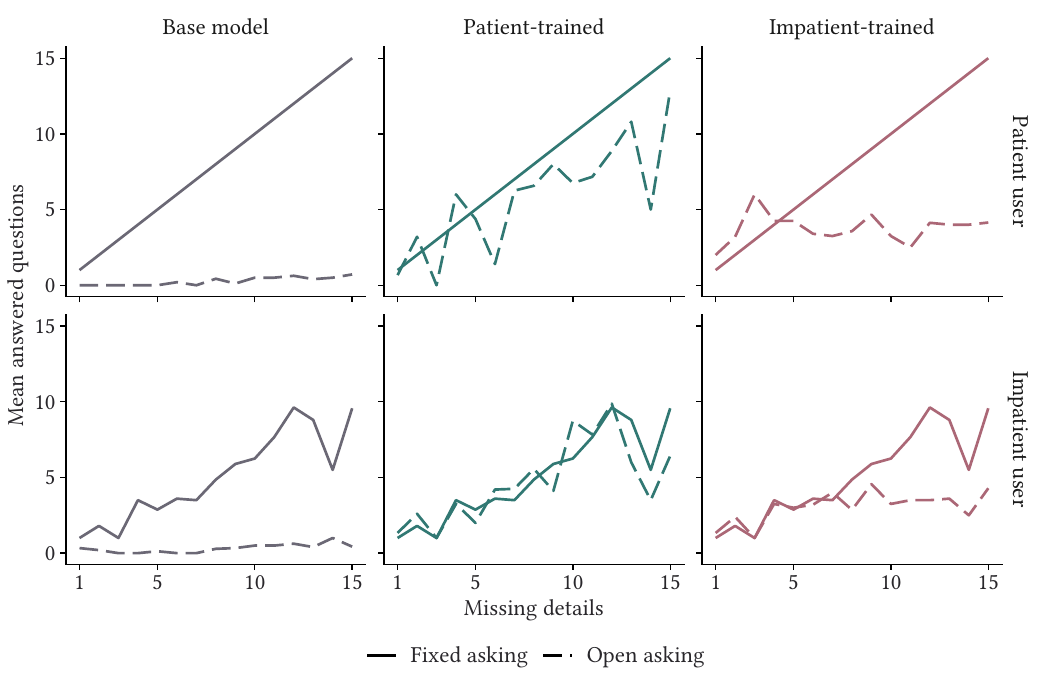}
    \caption{Mean answered-question count by number of redacted details, $k$, for each checkpoint and simulated-user type. Solid lines show fixed asking with budget $k$ and dashed lines show open asking with cap $2k$. Each point averages source-images with the relevant redaction count and impatient simulated users may terminate dialogue before the budget is exhausted.}
    \Description{Six panels comparing base, patient-trained, and impatient-trained checkpoints with patient and impatient users. Patient-user fixed-asking counts are diagonal, because they equal the number of missing details. Base open asking stays near zero. Trained open asking produces more answers, with variable counts that generally remain below fixed asking.}
    \label{fig:question_count_by_difficulty}
\end{figure*}

With patient simulated users, open asking reduces answered-question counts relative to fixed asking by 1.86 [0.73, 2.99] for patient-trained, $p_{\mathrm{BH}}=.002$, and 4.78 [3.78, 5.78] for impatient-trained, $p_{\mathrm{BH}}<.0001$ (\Cref{fig:question_count_by_difficulty}). With impatient users, the reduction is 2.27 [1.47, 3.07] for impatient-trained, $p_{\mathrm{BH}}<.0001$ and 0.45 for patient-trained which is not significant, $p_{\mathrm{BH}}=.404$.

Both trained checkpoints under-ask in open asking protocol, relative to the $k$-anchored fixed-asking baseline (\Cref{fig:question_count_by_difficulty}). Mean answered-question counts fall by 0.45--4.78 across user conditions and 3/4 reductions are significant ($p_{\mathrm{BH}}<.005$; patient-trained under impatient users is the exception, $p_{\mathrm{BH}}=.404$). This reduced elicitation accompanies lower observed recovery in every setting, with average declines of 7.2--13.3 percentage points.

Adjusting for question count predictably makes these recovery differences non-significant ($-5.5$--4.8 points, all $p_{\mathrm{BH}}\geq.120$), which is consistent with an effect from reduced elicitation. Together, these results suggest that models learn to ask when given discretion, but their stopping behavior leaves information unrecovered relative to the fixed-asking baseline calibrated to the number of redactions.

\subsubsection{How does user behavior affect performance?}
Interestingly, training with impatient users does not produce a clear advantage when evaluation users are also impatient. In these settings, the impatient-trained checkpoint has lower observed recovery than patient-trained under both fixed (65.9\% vs.\ 69.1\%) and open asking (57.7\% vs.\ 61.9\%). Impatient training also does not seem to eliminate sensitivity to early dialogue termination. Thus, these results provide no clear evidence that exposure to impatient users produces questioning strategies better suited to them, and we focus our human study on the patient-trained asker.

\subsubsection{How might this impact interaction burden?}
Looking at properties of the simulated interaction allows us to initially speculate about potential consequences for interaction burden; in other words, the cost (time, cognitive, or otherwise) that the user might incur to reap the benefits from these askers.

We find that training increases the length of exchanges even with equivalent question counts. Across fixed-asking comparisons, adjusted increases are 21.9--36.8 words per question and 7.2--10.3 words per answer, all $p_{\mathrm{BH}}<.0001$. Total dialogue consequently increases by 189.7--368.4 words. Recovery gains within a fixed number of questions may therefore be accompanied by greater reading and response demands, although we acknowledge word counts are a very imperfect proxy for this.

\begin{figure*}[!htb]
    \centering
    \includegraphics[width=\linewidth]{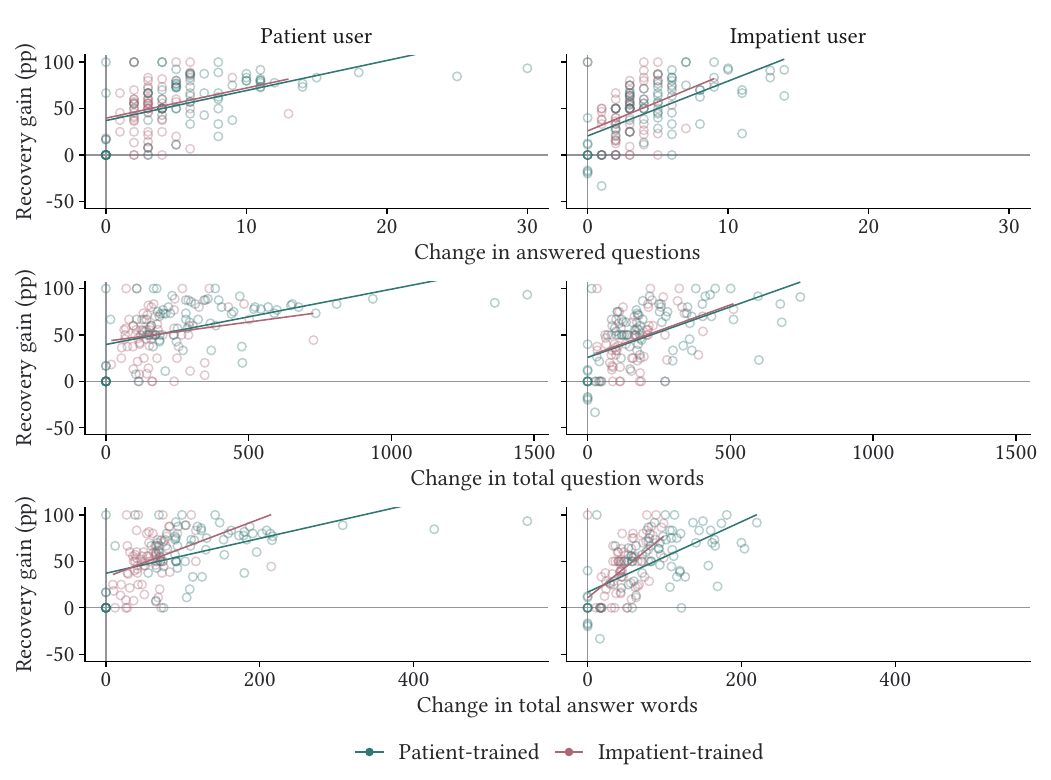}
    \caption{Recovery gain vs. change in interaction measure under open asking. Each point compares trained vs. base checkpoints on the same source image and simulated-user type; gray lines mark no change. Even low-change examples often produce non-zero recovery gain, suggesting the gains are not only driven by more complexity in interaction.}
    \Description{Six scatterplots arranged in three rows and two columns. Columns show patient and impatient simulated users. Every point is a trained-minus-base comparison for the same source image under open asking, patient-trained and impatient-trained. The vertical axis is recovery gain in percentage points, with ticks from -50 to 100. Horizontal axes show the change in answered questions in the top row, total question words in the middle row, and total answer words in the bottom row, with reference horizontal and vertical lines indicating zero change. Colored straight lines also summarize the association within each trained checkpoint. Most points lie above zero recovery gain and to the right of zero interaction change. All fitted lines slope upward, but points are widely dispersed around them. In the question-count panels, many examples with roughly 2–8 additional answers span recovery gains from near zero to 100 percentage points. Positive gains also occur at zero additional answers. A few negative recovery differences appear with impatient users near zero interaction change. Patient-trained comparisons extend furthest to the right.}
    \label{fig:training_gain_and_question_change}
\end{figure*}

In \Cref{fig:training_gain_and_question_change}, we plot changes in question count, question words, and answer words against recovery gains under open asking, paired at the source-image level. Gains generally accompany additional interaction, but positive gains also appear near zero change, suggesting improvements in how effectively the exchanges are used. Consistent with this interpretation, regression-adjusted training advantages remain positive after controlling for question count and question, answer, and output lengths, ranging from 6.0 to 17.2 percentage points (3/4 contrasts are significant, $p<.05$; with the exception of impatient-trained with impatient users, $p=.242$).

\subsection{Human Results}
\label{sec:human_results}

We filter the human data at 25 minutes (our preregistered 30-minute exclusion threshold plus a small grace window of 5 minutes for those who were close; a modest relaxation from our preregistration to account for our overestimation which affects only 3 participants), resulting in 76 valid participants' data (from 79 total) and 456 rounds in total. These retained rounds cover 77 reference images of which 71 appear in both conditions.

The median session duration was 52.9 minutes (IQR: 41.9--64.9). Across conditions, participants answered 1,757 total questions, with a median of three answers per round (IQR: 1--5) and a median round duration of 7.1 minutes (IQR: 4.8--10.3). Participants' initial prompts contained a median of 74 words (IQR: 43--121).

\subsubsection{Do simulated-trained policies improve results from interaction with people?}

\begin{figure*}[!htb]
    \centering
    \includegraphics[width=\linewidth]{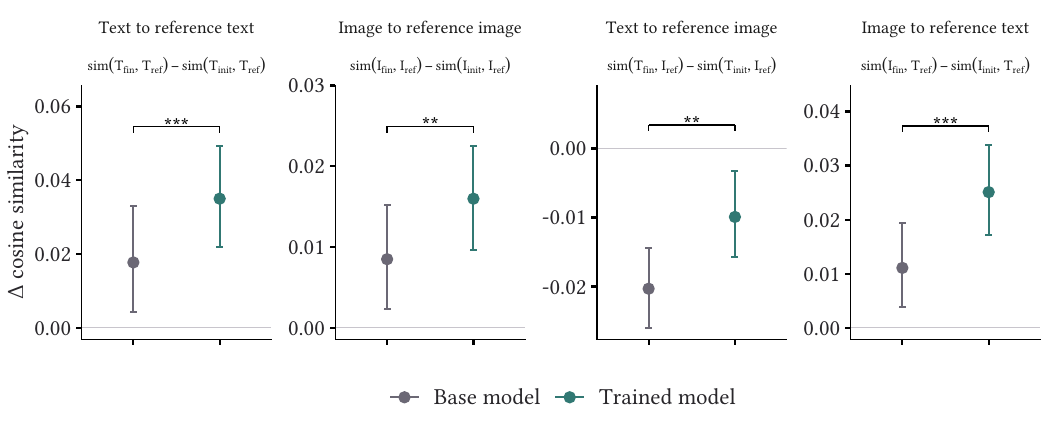}
    \caption{Fidelity improvement (cosine-similarity gains) in human interactions across text/image comparisons, using reference descriptions and images. Points show estimated marginal means; bars are 95\% participant-clustered regression confidence intervals. Brackets show trained-vs.-base contrasts adjusted for question count and source image, with participant-clustered standard errors. All contrasts favor the trained checkpoint. BH-adjusted significance: $^{**}p<.01$, $^{***}p<.001$.}
    \Description{Four side-by-side point-interval plots of human-study cosine-similarity gains. From left to right, panels compare text with reference text, image with reference image, text with reference image, and image with reference text. Gray points represent base and teal points the patient-trained assistant. Vertical bars are 95\% participant-clustered regression confidence intervals, and horizontal lines mark zero gain. The panels use separate vertical scales. The trained mean is higher in every panel, including a smaller decline in text--image similarity. Regression-adjusted trained-versus-base brackets have adjusted p < 0.001 for text--text and image--text, and < 0.01 for image--image and text--image.}
    \label{fig:cosine_gains}
\end{figure*}

The trained assistant results in significant reference-similarity gains across all text/image comparisons (\Cref{fig:cosine_gains}). After adjustment for question count and source image, advantages range from 0.0141--0.0337 cosine similarity, all $p_{\mathrm{BH}}<.005$. These include improvements in both the final description, with a text--text advantage of 0.0337 [0.0159, 0.0515], and the generated image, with an image--image advantage of 0.0141 [0.0045, 0.0237]. Thus, the learned policy improves final outcomes even when people formulate their own requests and answer the assistant's questions.

This effect persists across regression specifications; training coefficients are positive for every quality measure with and without question-count adjustment and source-image fixed effects, all $p<.02$ (\Cref{tab:quality_text_cosine_gain_regressions,tab:quality_image_cosine_gain_regressions,tab:quality_text_image_cosine_gain_regressions,tab:quality_image_text_cosine_gain_regressions}). Observed text--text gains average 0.0350 for the trained model vs. 0.0177 for base; image--image gains average 0.0160 vs. 0.0085, and image--text gains average 0.0251 vs. 0.0111. Text--image gains are negative in both conditions, which may reflect sensitivity in this cross-modal metric to changes in description length, specificity, and other such features. However, training still produces a relative advantage here.

\subsubsection{Do gains persist under comparable amounts of interaction?}
\begin{figure*}[!htb]
    \centering
    \includegraphics[width=\linewidth]{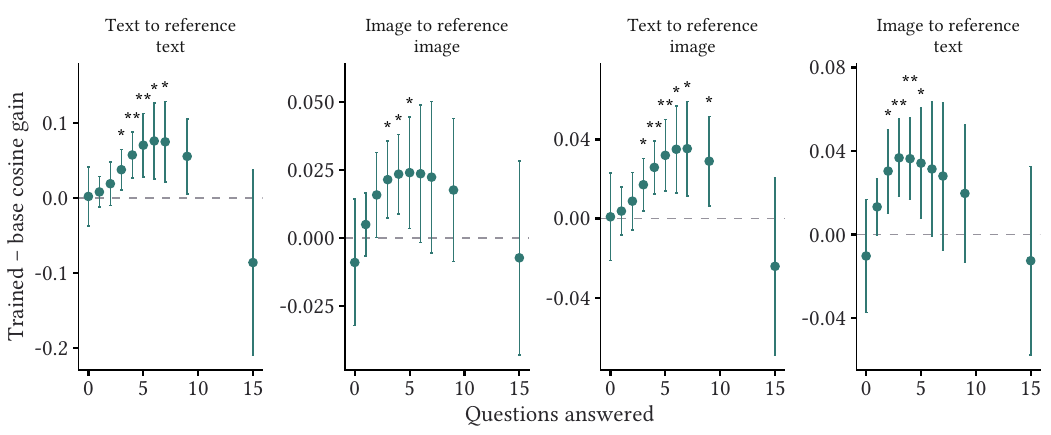}
    \caption{Adjusted differences (trained - base) in human reference-similarity gains by question count. Estimates come from a regression using a natural cubic spline, with source-image fixed effects. Bars are 95\% participant-clustered regression confidence intervals and the dashed line marks no difference. All shown count-by-outcome contrasts are BH-adjusted: $^{*}p<.05$, $^{**}p<.01$.}
    \Description{Four side-by-side point-and-interval plots of the adjusted trained-minus-base difference in human-study cosine-similarity gain. Panels show text--text, image--image, text--image, and image--text comparisons. The horizontal axis is answered-question count, spanning 0 to 15, with estimates at counts observed in both conditions. Points are contrasts from the training-by-natural-cubic-spline model; vertical bars are 95\% participant-clustered regression intervals. Dashed horizontal line marks no difference. Vertical scales differ across panels. Estimates at zero and one answer are near zero and have intervals crossing zero. At three to five answers, all four panels show positive contrasts with adjusted p < 0.05. The fitted profiles generally rise through the first few counts and then flatten or decline. Text--text and text--image estimates remain positive across more of the middle counts. At 15 answers, all point estimates are negative, but their intervals are much wider and cross zero. No estimate at that count has a significance mark.}
    \label{fig:training_by_question_count}
\end{figure*}

Our primary comparisons account for question count, but effectively assume a common training advantage across counts. In \Cref{fig:training_by_question_count}, we relax this assumption to study how advantage varies with the number of answers participants actually provide. Since estimates at individual counts are noisy, we interact training condition with a natural cubic spline of question count, keeping intact the source-image fixed effects and participant-clustered standard errors. This allows training advantage to vary smoothly with question count, sharing information across nearby counts, but also not requiring a more restrictive linear or monotonic relationship.

Training advantages are positive and significant across all four quality measures at 3--5 answered questions, all $p_{\mathrm{BH}}<.05$. At four answers, for instance, the text--text advantage is 0.057 [0.026, 0.088] and the image--image advantage is 0.023 [0.009, 0.038]. Evidence at other counts varies by measure but importantly, no comparison is significant at zero or one answer, which is consistent with the interaction being what matters in practice. The estimate at high count (15) has limited support and thus very wide CIs; we do not draw any conclusions from this.

\subsubsection{How does this affect the realized burden on actual users?}
\begin{figure*}[!htb]
    \centering
    \includegraphics[width=\linewidth]{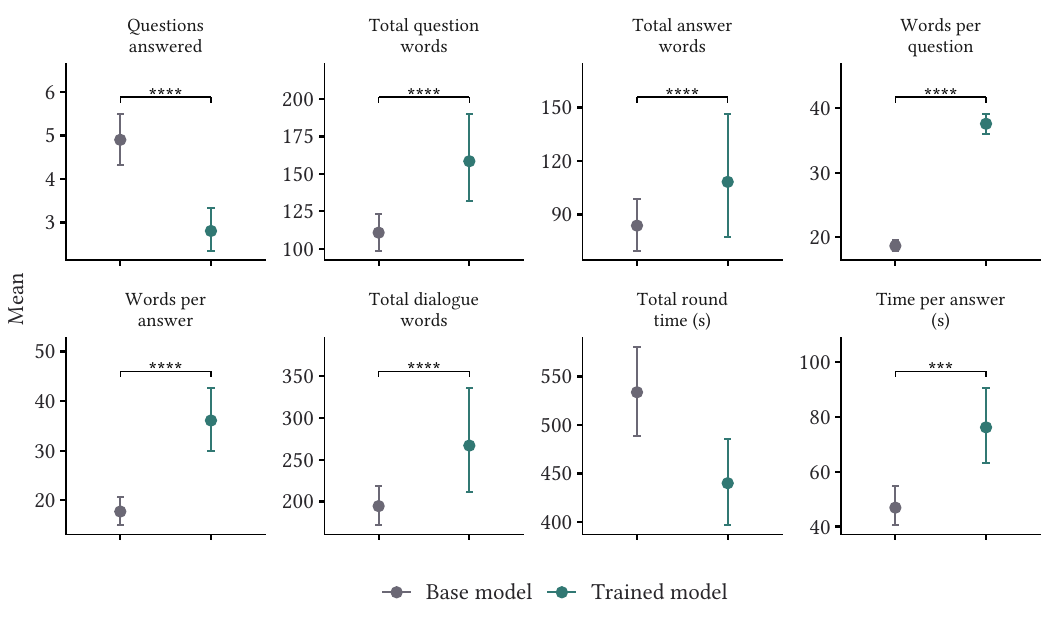}
    \caption{Interaction measures from the human study. Points show estimated marginal means; bars are 95\% participant-clustered regression confidence intervals. Brackets show trained-vs.-base regression-adjusted contrasts and all tests use participant-clustered standard errors. BH-adjusted significance: $^{*}p<.05$, $^{***}p<.001$, $^{****}p<.0001$.}
    \Description{Eight point-interval plots arranged in two rows of four, base vs. patient-trained assistant; bars are 95\% participant-clustered regression confidence intervals. The top row shows questions answered, total question words, total answer words, and words per question. The bottom row shows words per answer, total dialogue words, total round time, and time per answer. Each panel has its own vertical scale. Adjusted means are 5.03 questions for base and 2.76 for trained. Question length increases from approximately 18 to 39 words, and answer length from 19 to 36 words. Adjusted total dialogue increases from 157 to 347 words per round; both question and answer word totals are higher with the trained assistant. Adjusted time per answer rises from approximately 49 to 72 seconds, while adjusted total round duration is similar, at approximately 489 and 491 seconds. Adjusted comparison brackets mark p < 0.0001 for count and word measures and below 0.001 for time per answer. Total round time difference is not significant.}
    \label{fig:survey_interaction_burden}
\end{figure*}

We find that participants answer \textit{fewer} questions with the trained assistant despite getting improved results, as shown in \Cref{fig:survey_interaction_burden} (adjusted difference is $-2.27$ answers [$-2.82$, $-1.72$], $p_{\mathrm{BH}}<.0001$; observed means are 2.81 versus 4.90). This differs from the simulated open asking setting, wherein training increased questioning over the base model which rarely asks.

The exchanges themselves are longer; after adjustment, questions are on average 20.9 words longer [19.1, 22.7] and answers are 16.8 words longer [11.4, 22.2], both $p_{\mathrm{BH}}<.0001$. At common question count, we estimate 189.8 additional total dialogue words with the trained assistant [134.6, 245.1], $p_{\mathrm{BH}}<.0001$. However, since the observed conversations are marked by fewer interactions with the trained model, the realized difference is smaller at 72.3 words, with means of 266.8 versus 194.5. Observed rounds are also faster with the trained assistant overall, averaging 439.9 seconds versus 533.7 for base.

The training therefore appears to change how the work is distributed, i.e. participants respond fewer times, but read and write more text when doing so. Paired with the improved results, we hypothesize that this may reflect fewer, more precise questions that elicit more consequential responses from users (although we caution against treating these results as strong evidence of this interpretation). We note that these measures do not, of course, constitute a reduction in subjective cognitive load.

\subsubsection{Does better clarification come at a cost?}
\begin{figure*}[!htb]
    \centering
    \includegraphics[width=\linewidth]{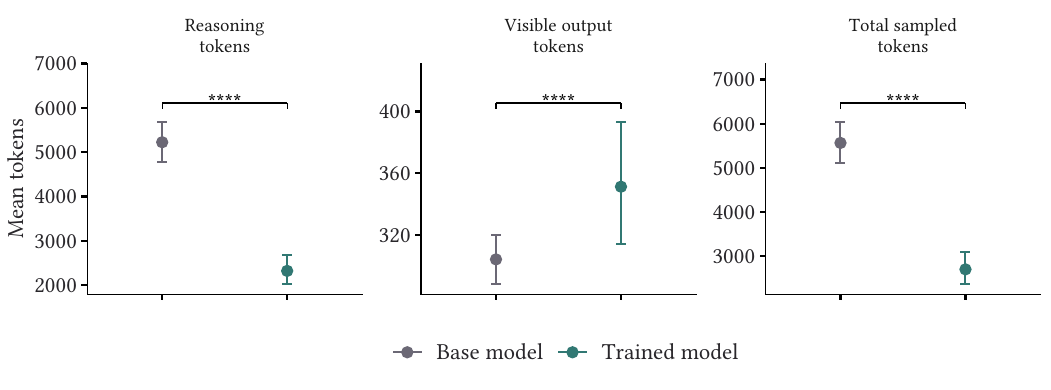}
    \caption{Model-side token usage per human-study round. Points show estimated marginal means; bars are 95\% participant-clustered regression confidence intervals. Brackets show regression-adjusted trained-vs.-base contrasts with participant-clustered standard errors. Results indicate that the trained model uses fewer reasoning and total sampled tokens but more visible-output tokens. All contrasts have $p_{\mathrm{BH}}<.0001$ ($^{****}$).}
    \Description{Three side-by-side point-interval plots of mean assistant tokens per human-study round. Panels show reasoning tokens, visible-output tokens, and total sampled tokens for base vs. the patient-trained assistant, with 95\% participant-clustered regression confidence intervals. Each panel uses its own vertical scale. Adjusted reasoning usage falls from approximately 4,597 tokens with base to 3,046 with trained. Adjusted visible-output usage increases from 273 to 403 tokens. Adjusted total sampled usage falls from 4,898 to 3,478 tokens. The reasoning reduction exceeds the visible-output increase. Confidence intervals are separated between conditions in all three panels. Brackets show regression-adjusted contrasts, all with adjusted p < 0.0001.}
    \label{fig:token_usage}
\end{figure*}

We find that training improves reference similarity while \textit{reducing} token usage (\Cref{fig:token_usage}). Even after accounting for question count and source image, the trained assistant uses 1,420 fewer tokens per round [1,146, 1,695], $p_{\mathrm{BH}}<.0001$. This reduction therefore persists when comparing conversations with the same number of answers.

The change is most apparent in internal reasoning tokens, which decrease by 1,551 per round, while visible-output tokens increase by 131; both $p_{\mathrm{BH}}<.0001$. Observed total usage averages 2,698 tokens with trained versus 5,563 with base. The assistant thus produces effective user-facing exchanges with less internal reasoning, suggesting that learning useful clarification behavior can also reduce the inference-time computation needed to support it.

\subsubsection{How do the model-generated questions and human-given answers change after training?}
To investigate this, we use the \textit{Matryoshka} interpretability technique~\cite{cherep2026visualpersuasion,sharif2026improving} to recursively summarize over source-image-paired question sets and answer sets from control vs. treatment conditions. The summaries suggest differences in question format and response style, including e.g. fewer binary/polar questions and less hedging and speculative answers. See \Cref{app:autointerp} for more details on the procedure and examples.

\section{Discussion}
Overall, our results indicate that assistants can learn to clarify with real users through simulated interaction. A broader design question we might then ask is what this capability should accomplish for the user. In other words, why should we want this, and what might its impacts be? We argue that productive clarification should render users' knowledge, judgments, and preferences more valuable in the long term, and develop this and related implications here.

\subsection{Eliciting authorship through participation}
In our study, learned clarification changed the trade-offs of delegation. Participants using the trained assistant answered fewer questions but wrote longer responses while obtaining better results. One way to look at this is that clarifying before proceeding gives the user an opportunity to determine part of the outcome. Such participation may support authorship by making the user's judgments more operative in an artifact that the model produces. 

Kreminski~\cite{kreminski2024dearth}, for instance, describes the \textit{dearth of the author} as a condition in which generated output contains many decisions but the amount of human expressive intent is not commensurate with it, and proposes that assistants could invert this by drawing out intentions and helping users refine them~\cite[cf.][]{kreminski2024intent}. Learning to clarify offers a concrete mechanism for this, since questions can return otherwise automated decisions to users. Such exchanges might also encourage users to examine what they want before accepting assistants' responses, conceptually in line with evidence from cognitive forcing functions and overreliance~\cite{buccinca2021trust}. Relatedly, human writers have been shown to sometimes even revise their work to accommodate machine suggestions~\cite{singh2023hide}. Effective clarification could help assistants discover constraints their suggestions should satisfy before producing them, facilitating a more reciprocal adaptation between writer and assistant. Questions can also draw attention to overlooked details or help users recognize what needs to be stated explicitly~\cite{singh2024figura11y}. Arguably, these stakes extend to delegated action broadly: legal questions about responsibility and intent under AI use have the potential to make such opportunities to exercise intent even more consequential~\cite{ayres2024law}.

\subsection{Without clarification, missing intent may leave room for undesirable influence}
Underspecified requests leave assistants reliant on their priors, and these can in turn be undesirable at times. For instance, implicit perceptual priors of multimodal language models appear to even include negative stereotypes~\cite{cherep2026probing}. Under uncertainty, LLM-based agents might also be more susceptible to external influences such as nudges and visual presentation~\cite{cherep2026visualpersuasion,cherep2026ai}. In one study~\cite{cherep2025framework}, matching product ratings and prices left many LLMs strongly susceptible to nudges, but explicit user profiles suppressed these effects. This suggests that resolving uncertainty about user preferences for AI assistants may also serve to affect which other signals govern their decisions. Future work might thus investigate whether improved clarification also constrains sensitivity to external influence in delegated work.

\subsection{Learning to establish common ground}
Users cannot be realistically expected to anticipate every detail their assistant will need in advance. In our task, participants could freely and limitlessly inspect the reference image throughout the interaction, and yet still omitted details, enough that an improved clarification policy was able to help significantly. As Vafa et al. note~\cite{vafa2026s}, model-producible results may not even be reachable by human prompters. Additionally, prompting even image generators (where there is readily-apparent visual evidence) can be quite unintuitive, involving specialized practices~\cite{oppenlaender2024taxonomy,oppenlaender2025prompting}. Questions could thus help draw attention to something overlooked, or recognize the need to state some aspect of the task specification explicitly. Clarification can therefore help users decide what to communicate, as well as help assistants interpret what they say.

We could look at this as part of establishing \textit{common ground}, wherein participants in a conversation work together to establish sufficient understanding for the task at hand~\cite{clark1991grounding}. This is arguably a broader ambition in the history of HCI and AI~\cite{brennan1998grounding,grosz1996collaborative}. As illustrated by Picard, now more than three decades ago~\cite{picard1995computer}:
\begin{discussionquote}
``When I say `city scene,'\,'' said Arthur,
``I want a lot of skyscrapers, a skyline; the
prototypical city---you know, cars, smog.''

``When I think of a city,'' said Beth, ``I think
of beautiful cities---with nicely landscaped
green areas around each building, and colorful
people milling about.''

{\normalfont\footnotesize
\raggedleft --- Picard, 1995~\cite{picard1995computer}\par}
\end{discussionquote}

Here, Arthur and Beth use similar words to describe different implicit expectations; this is being revealed perhaps in response to a clarification question. Our results suggest that AI assistants can learn to support this process through interaction, therefore reducing the need for users to anticipate, in their initial requests, all the ways assistants might misunderstand them.

\subsection{Mutually improving simulation and human-AI interaction}
The transfer effects we observe suggest that simulated users can support learning useful interaction strategies even when they imperfectly reproduce human behavior. Prominent recent work has demonstrated the potential of such simulations~\cite{park2023generative,horton2023large,manning2026general}. However, recent findings also document systematic distortions in the behavior of such simulations~\cite{peng2026digital}. Such distortions make human evaluation necessary, but can also give it a constructive role in bettering the simulations: experiments can help identify which aspects of simulated behavior need refinement to support the interactions we wish to learn.

Our results already suggest one such refinement. Participants answered fewer questions with the trained assistant, a pattern our heuristic model of impatience was not designed to capture. We hypothesize that users may proceed to generation once they feel they have communicated enough useful information, while continuing to interact when important information remains uncommunicated. User patience may thus depend on the perceived value of further information exchange. Future simulations could use privileged knowledge of uncommunicated factors to model this dependence. This would allow human participation to inform many simulated interactions, reducing the burden for users to supply the training trajectories themselves.

\section{Limitations}
The value-of-information interpretation we take depends on how faithfully our objective represents what users value. We operationalize utility as recovery of initially unspecified details, which is one specific choice, and evaluate generated outputs through embedding similarity to a reference. Though these measures help to make information acquisition tractable, we acknowledge that they cannot establish the relative value of each recovered detail or that greater scores improve subjective satisfaction. We use a heuristic LLM-as-judge importance-weighting scheme to help mitigate this and test robustness to nonuniform value, but we believe future efforts will need to test other operationalizations which are sensitive to differing users' and circumstances' specific goals.

Importantly, the reference reconstruction task we use with participants gives them a stable target that they can inspect throughout the interaction. This makes the whole study tractable, but we acknowledge that delegation may often involve uncertain or evolving preferences, degrees of tolerable variation, and decisions that users explicitly want their assistants to make. Extending our conclusions to such settings requires further experimentation, and we believe this is necessary.

More broadly, the transfer effects we document are bounded by the single task domain, single assistant model, and a comparison with its untrained counterpart. Our human evaluation provides an independent test of interaction behavior, but there is a wide space of different simulators, generators, or alternative clarification methods one could test; we see our work as the first step on this path, and almost surely not the global optimum. Testing such changes is necessary to determine which learned behaviors generalize beyond these conditions into the broader space of possible tasks and interactions involving clarification.

\section{Conclusion}
Even as more task execution work is automated, we have reason to expect that task \textit{specification} will remain challenging. We typically cannot declare in advance everything that might matter to us, and our omissions are not random: we often leave out what seems obvious when making a request, even though it may admit several reasonable interpretations from the assistant that receives it. The task can therefore be ill-posed even when the assistant possesses the capability to execute any of those interpretations. Clarification policies help re-center interaction, giving it the role of establishing the desired interpretations. This suggests an ambition well beyond completing individual requests: developing systems that become better at making good use of what their users know. We invite the community to build on our framework toward assistants that make human knowledge and judgment \textit{more}, not less, consequential as their capabilities grow.


\begin{acks}
We received funding from SK Telecom with MIT's Generative AI Impact Consortium (MGAIC). Research reported in this publication was supported by an Amazon Research Award, Fall 2024. Experiments conducted in this paper were generously supported via API credits provided by Thinking Machines and Google. MC is supported by a fellowship from ``la Caixa'' Foundation (ID 100010434) with code LCF/BQ/EU23/12010079. The authors acknowledge the MIT Office of Research Computing and Data for providing high performance computing resources that have contributed to the research results reported within this paper.
\end{acks}


\bibliographystyle{ACM-Reference-Format}
\bibliography{main}

\newpage

\appendix
\crefalias{section}{appendix}
\crefalias{subsection}{subappendix}
\crefalias{subsubsection}{subsubappendix}
\FloatBarrier

\section{Examples}
\label{app:examples}

\begin{figure}[!htb]
  \centering
  \includegraphics[width=0.86\linewidth,keepaspectratio]{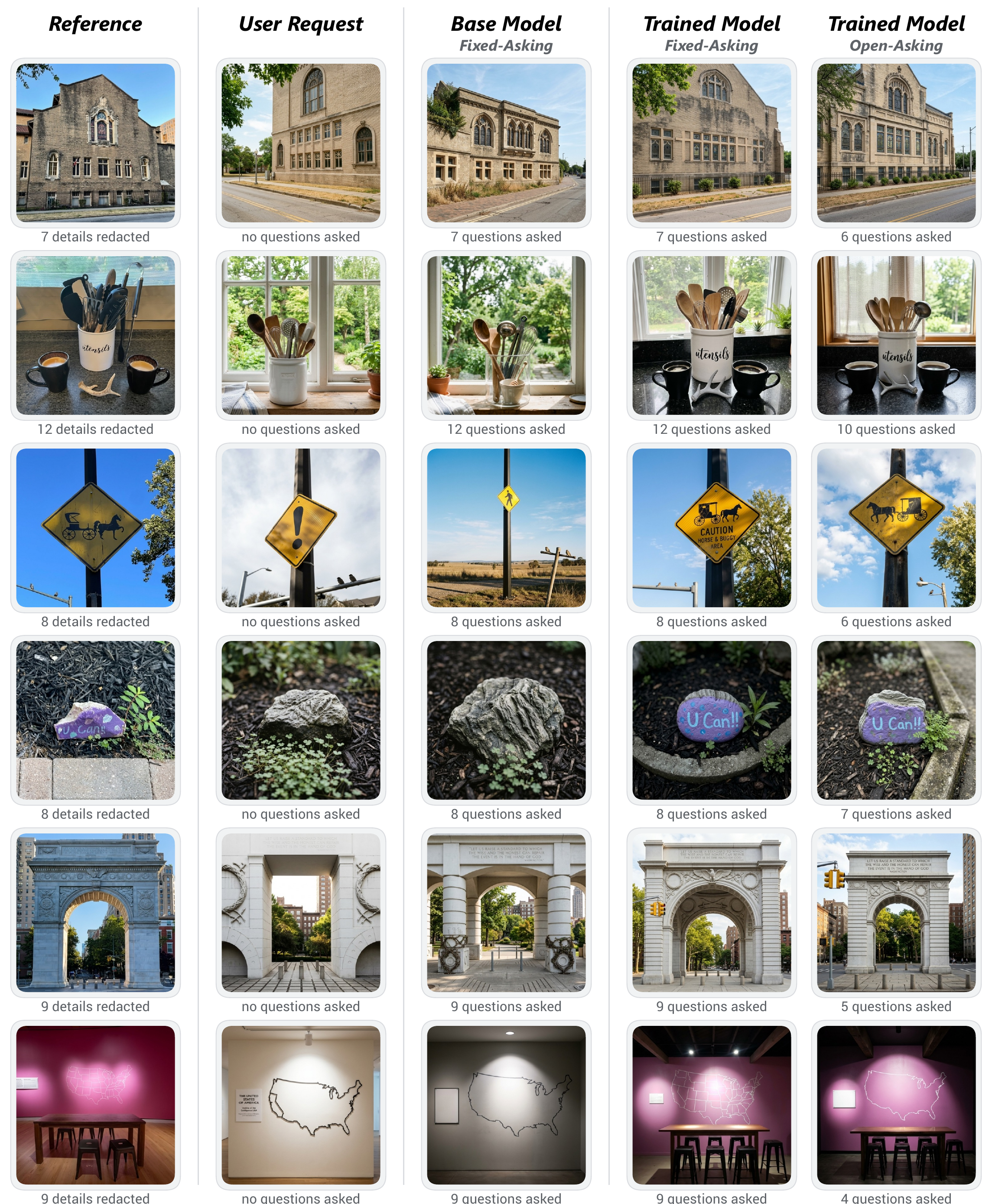}
  \caption{Generated images for six test instances, next to the reference photograph and the image from the request alone.}
  \Description{Six rows of five images. In each row the reference photograph is followed by the request-only image, the base-model image and the two trained-assistant images. The request-only and base-model images miss the hidden subjects, such as the horse and buggy on a caution sign, the painted rock and the pink wall, and the two trained-assistant images bring them back.}
  \label{fig:examples}
\end{figure}

\begin{figure}[p]
  \centering
  \includegraphics[width=\linewidth,height=0.92\textheight,keepaspectratio]{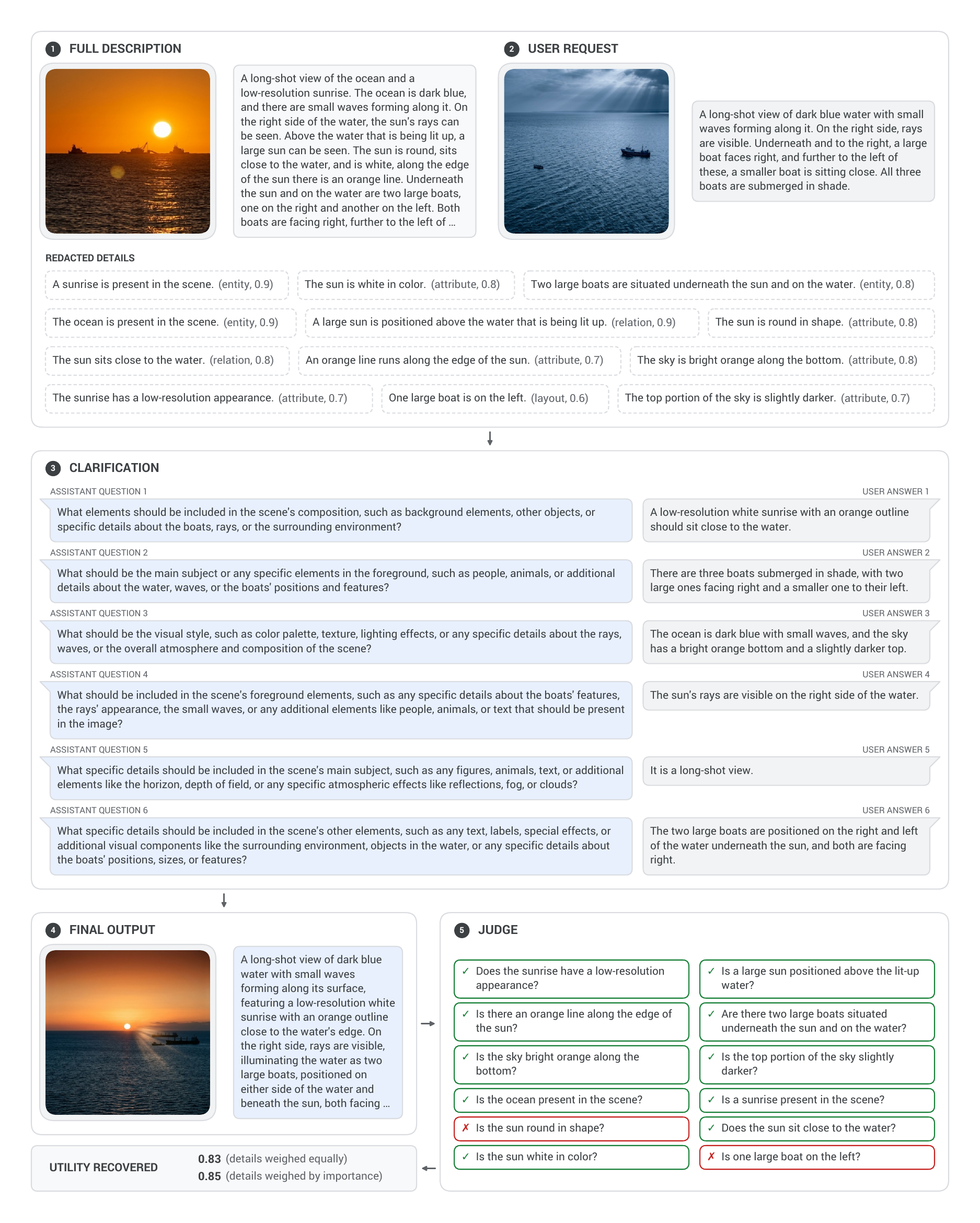}
  \caption{One test episode of the patient-trained assistant in open-asking mode, from the reference image to the judge's verdicts.}
  \Description{A one-page walkthrough of one episode. The top shows a sunrise photograph over the ocean with two boats and its full description, then the request, which omits the sunrise, and twelve redacted details. The middle shows six questions from the assistant, each answered by the user with one or two details. The bottom shows the generated image, which now contains the sunrise and the boats, and twelve judge checks, ten marked satisfied and two not.}
  \label{fig:example_instance}
\end{figure}

\Cref{fig:examples} shows example images from trained and untrained models, next to the reference and user request images. \Cref{fig:example_instance} shows one complete episode of the trained assistant, from the request to the judge's verdicts.

\FloatBarrier
\section{Instance Synthesis}
\label{app:synthesis}

Every instance is built from a single DOCCI description~\cite{onoe2024docci}, which passes through the decomposition model and then the redaction model. Both are Gemini 3.5 Flash-Lite, prompted as shown in \Cref{app:prompts}. All random choices are seeded by the source, so that the same source always yields the same instance. \Cref{tab:example_instance} shows the resulting request and hidden details for one test instance (the caution sign of \Cref{fig:examples}).

\paragraph{Sources.} We take the first 2,000 DOCCI descriptions of at least 400 characters. Each description is paired with the photograph it describes, which serves as the reference image.

\paragraph{Decomposition.} The decomposition model reads the full description $s^\star$ and returns the details $\mathcal{D}$ as structured output. For every detail, it produces a self-contained sentence, an importance weight in $[0, 1]$, and the binary check that the judge later answers. We drop sources whose decomposition yields fewer than $K + 5 = 20$ details, so that every instance keeps at least five details after hiding up to $K$ of them. This removes 1,521 of the 2,000 sources and leaves 479 instances with between 20 and 45 details each.

\paragraph{Redaction.} For each instance, we first draw the hidden count $k \sim \mathrm{Uniform}\{1, \dots, \min(K, m - 1)\}$ and then draw $k$ details one at a time without replacement, with probability proportional to $\max(w_i, 0.05)$. Important details are thus hidden more often, but no detail is unreachable. Across the 479 instances, this hides 3,791 details in total (7.9 per instance), spread almost evenly over the counts 1 to 15 with between 23 and 39 instances per count.

\paragraph{Request.} The redaction model receives the full description together with the texts of the hidden details, and writes the request $x$: one to three sentences in the register of a real user request that state every kept detail and none of the hidden ones. In \Cref{tab:example_instance}, for instance, the request states the 14 kept details of its source and none of the 8 hidden ones. An instance whose request came back empty would have been dropped, but this did not occur.

\paragraph{Splits.} We assign sources to the training, validation, and test splits with shares of 0.6, 0.2, and 0.2, using a seeded hash of the source id. This yields 290, 111, and 78 instances, respectively.

\begin{table}[!htb]
  \caption{The request and the hidden details of one test instance.}
  \label{tab:example_instance}
  \small
  \begin{tabular}{p{\linewidth}}
    \toprule
    \textbf{Request} \\
    \midrule
    An outdoor medium shot looks up at a yellow, diamond-shaped caution sign attached to a large, dark vertical pole stretching from top to bottom. The sign displays shadows and light reflections. A light pole is in the bottom corner of the shot with two birds sitting on it during daytime. \\
    \bottomrule
  \end{tabular}

  \vspace{0.6em}
  \begin{tabular}{p{0.41\linewidth}lp{0.41\linewidth}}
    \toprule
    Hidden detail ($\delta_i$) & Importance ($w_i$) & Check ($\chi_i$) \\
    \midrule
    The caution sign is weathered. & 0.6 & Is the caution sign weathered? \\
    The caution sign has a black-line outline. & 0.7 & Does the caution sign have a black-line outline? \\
    The caution sign features a black horse and buggy. & 0.9 & Does the caution sign feature a black horse and buggy? \\
    The black horse and buggy on the sign are facing the right. & 0.7 & Are the horse and buggy facing the right? \\
    The background consists of a blue sky. & 0.7 & Is the background a blue sky? \\
    A tree is present in the scene. & 0.5 & Is a tree present in the scene? \\
    The light pole extends to the left in the bottom corner of the shot. & 0.7 & Does the light pole extend to the left in the bottom corner of the shot? \\
    The two birds are sitting apart from each other on the light pole. & 0.6 & Are the two birds sitting apart on the light pole? \\
    \bottomrule
  \end{tabular}
\end{table}

\FloatBarrier
\section{Training and Evaluation Details}
\label{app:training}

This appendix provides the implementation details behind \Cref{sec:training,sec:analysis}. We describe how the assistant and the environment models are called, how the reward is computed, the training hyperparameters (\Cref{tab:hyperparameters}), how we selected checkpoints, and the settings used for evaluation and for rendering images.

\paragraph{Assistant calls.} The assistant is Qwen3-8B, which we serve and train through the Tinker API using the model's recommended chat template with its reasoning mode on. Each call samples at most 4,096 tokens, shared between the reasoning and the visible text. If a completion hits this cap or leaves a reasoning tag in its visible text, we discard it and sample once more. Only the visible text enters the history and reaches the user and the judge, while the reasoning is stored with the episode.

\paragraph{Environment calls.} The decomposition model, the redaction model, the user simulator, and the judge each make one structured call to Gemini 3.5 Flash-Lite. The system message consists of the prompt in \Cref{app:prompts} followed by the list of input fields the call receives, the user message carries the field values under the same names, and the response is JSON that we validate against the output schema (whose fields are listed with each prompt). A failed call is retried up to three times with exponential backoff. In our runs, the simulator ran out of retries in only ten of the roughly 150,000 training episodes.

\paragraph{Judge.} For every hidden detail, the judge receives its check $\chi_i$ together with the composed description $y$ alone, and returns whether $y$ states or clearly entails the fact. The $k$ checks of an episode run concurrently. We then compute $\urec$ both under the importance weights (floored at 0.05, as in the redaction sampling) and under uniform weights.

\paragraph{Importance and uniform weights.} Every evaluation computes $\urec$ under both weightings, and the two move closely together. Across the 15 evaluations of the test split in \Cref{sec:analysis} (one per combination of assistant, asking mode, and user, 1,170 episodes in total), the mean under importance weights exceeds the mean under uniform weights by 0.007 on average and by at most 0.014. Within an evaluation, the two values of an episode correlate at 0.985 or above (median 0.994), and ranking the 15 evaluations by either mean gives the same order. The validation split gives the same picture, with a gap of at most 0.012 and again an identical ranking. The importance-weighted value runs slightly higher because the assistants recover the heavier hidden details somewhat more often than the lighter ones. The choice of weighting therefore changes no comparison between assistants, modes, or users.

\paragraph{Policy gradient.} \Cref{tab:hyperparameters} lists the settings shared by both training runs, which differ only in the user simulator. Each iteration draws $M = 8$ training instances with a seeded sampler, so that a resumed run continues the same schedule, and collects $g = 32$ episodes per instance. Under the impatient user, the patience $p$ is drawn once per instance and iteration from the seed, the iteration, and the instance id, so that all 32 rollouts of a group face the same patience and the luck of the draw does not enter the group-centered advantage. Every rollout is judged, its reward is $\urec$ under uniform weights, and its advantage is $A_j = r_j - \frac{1}{g} \sum_{j'=1}^{g} r_{j'}$, the reward minus the mean of its group. The group mean serves as a baseline: it estimates the reward the current policy earns on the instance, including the share of hidden details the assistant tends to state from $x$ alone, so that each rollout is scored against the other rollouts on the same instance rather than against rollouts on easier or harder instances. Which of those details a particular rollout happens to state, and how well the simulator answers it, still vary from rollout to rollout. Centering does not remove this variation, which stays in $A_j$ as zero-mean noise, but subtracting a baseline does not change the direction of the expected gradient and lowers its variance~\cite{williams1992simple}, so the update favors questions that recovered more than the group average.

We then take one step on the LoRA parameters $\theta$ (rank 32)~\cite{hu2022lora} that descends the importance-sampled policy-gradient loss
\begin{equation}
\label{eq:loss}
\mathcal{L}(\theta) = - \sum_{j} A_j \sum_{t \in \mathcal{T}_j} \frac{\pi_\theta(w_t \mid c_t)}{\pi_{\theta_{\mathrm{old}}}(w_t \mid c_t)},
\end{equation}
where $j$ runs over the $Mg$ episodes of the iteration, $\mathcal{T}_j$ over every token the assistant sampled in episode $j$ (in its questions and its description, reasoning included), $w_t$ is such a token, $c_t$ the prompt and the tokens before it, and $\theta_{\mathrm{old}}$ are the weights the episodes were sampled from. Prompt tokens, which include the user's answers, carry no weight. Because the policy is updated once per batch, $\theta = \theta_{\mathrm{old}}$ when the gradient is taken and every ratio equals one, so the update reduces to the policy gradient $\sum_j A_j \sum_{t \in \mathcal{T}_j} \nabla_\theta \log \pi_\theta(w_t \mid c_t)$ (i.e., REINFORCE with a group baseline), and the ratio only corrects for numerical differences between the sampling and the training weights. The loss has no clipping of the ratio as in PPO~\cite{schulman2017proximal}, no KL penalty toward the base model, no entropy bonus, and no length normalization: it sums over tokens, so an episode's advantage is applied once for each token it sampled. The step is an Adam step with learning rate $4 \times 10^{-5}$ and otherwise the defaults of the Tinker API ($\beta_1 = 0.9$, $\beta_2 = 0.95$, $\epsilon = 10^{-12}$, no weight decay, no gradient clipping). A checkpoint is saved every five iterations.

\begin{table}[htb]
  \caption{Training settings shared by the patient-trained and the impatient-trained assistant.}
  \label{tab:hyperparameters}
  \small
  \begin{tabular}{ll}
    \toprule
    Base model & Qwen3-8B, reasoning mode on \\
    Adapter & LoRA, rank 32 \\
    Optimizer & Adam, learning rate $4 \times 10^{-5}$, $\beta_1 = 0.9$, $\beta_2 = 0.95$, $\epsilon = 10^{-12}$ \\
    Regularization & no weight decay, no gradient clipping \\
    Instances per iteration $M$ & 8 \\
    Rollouts per instance $g$ & 32 \\
    Asking mode in training & fixed-asking, $B = k$ \\
    Reward $r(\tau)$ & $\urec$ under uniform weights \\
    Advantage $A_j$ & reward minus the group mean, no normalization \\
    Loss & importance-sampled policy gradient (\Cref{eq:loss}), one step per batch \\
    Omitted terms & no ratio clipping, no KL penalty, no entropy bonus, no length normalization \\
    Credited tokens & every sampled token of every assistant call, reasoning included \\
    Sampling cap & 4,096 tokens per call \\
    Iterations & 283 (patient), 300 (impatient) \\
    Checkpoints & every 5 iterations \\
    Random seed & 42 \\
    \bottomrule
  \end{tabular}
\end{table}

\begin{figure*}[!htb]
    \centering
    \includegraphics[width=\linewidth]{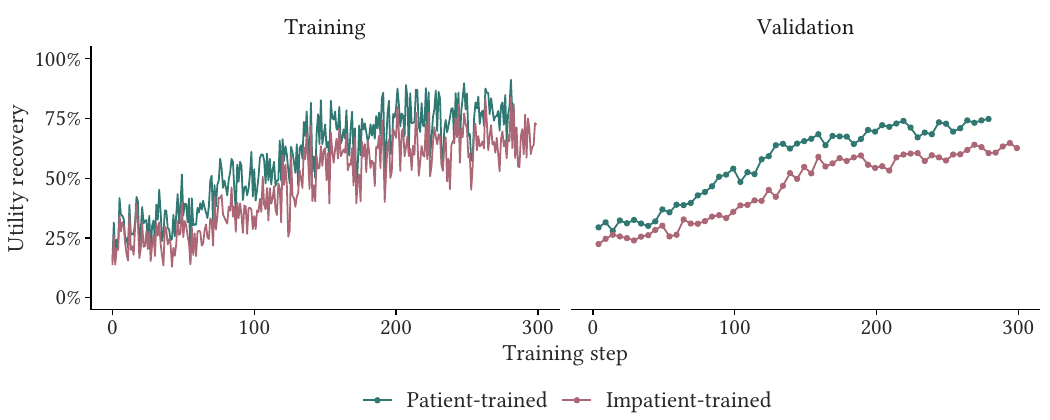}
    \caption{Mean training reward per iteration and validation recovery of the checkpoints saved every five iterations, for each run against the user it was trained with.}
    \Description{Two panels with the training step from 0 to 300 on the horizontal axis and utility recovery from 0 to 100 percent on the vertical axis, one line per run. The training panel is noisy from iteration to iteration, rising from about 20 percent to about 70 percent by step 150 and then staying flat. The validation panel is smoother, rising from about 30 percent to about 75 percent for the patient-trained run and from about 22 percent to about 63 percent for the impatient-trained run, with the patient-trained line above the impatient-trained line throughout.}
    \label{fig:training_curves}
\end{figure*}

\paragraph{Checkpoint selection.} We evaluate every checkpoint in fixed-asking mode on the validation split, against the user it was trained with (\Cref{fig:training_curves}). The patient run was stopped at iteration 283 when the reasoning length started to grow (about 1,300 tokens per call at iteration 275 and 2,200 at iteration 282). We take iteration 219, the earliest checkpoint on the plateau, which reaches a validation $\urec$ of 0.740 under importance weights and 0.732 under uniform weights at 7.28 answered questions (i.e., the mean hidden count of the validation split, since the patient user answers all $k$ questions). The impatient run trained for the full 300 iterations. Its iteration 294 scores best under the impatient user, with 0.647 and 0.643 at 4.75 answered questions, and reaches 0.717 and 0.711 at 7.28 answered questions under the patient user.

\paragraph{Evaluation.} An evaluation runs one assistant over one split using the prompts, judge, and instances of training. Under the impatient user, the patience $p$ is drawn per instance from the seed and the instance id, so that it is identical for every assistant and mode and comparisons across assistants are paired.

\paragraph{Images and embeddings.} We render both the request and the composed description of every episode once with Gemini 3.1 Flash Image at $512 \times 512$ pixels (aspect ratio 1:1), with the instruction of \Cref{app:prompts} appended and up to three retries. Similarities are cosine similarities between Gemini Embedding 2~\cite{shanbhogue2026gemini} embeddings of these texts and images and those of the DOCCI description and photograph, within and across modalities.

\FloatBarrier
\section{User Simulator Audit}
\label{app:audit}

\begin{table}[!htb]
  \caption{Audit of the simulator's answers in fixed-asking mode under the patient user.}
  \label{tab:audit}
  \small
  \begin{tabular}{lrr}
    \toprule
    & Patient-trained & Impatient-trained \\
    \midrule
    Hidden details per answer & 1.06 & 0.74 \\
    Answers with two or more hidden details & 31\% & 20\% \\
    Stated details asked for specifically & 18\% & 15\% \\
    Stated details asked for by topic & 81\% & 83\% \\
    Stated details volunteered & 1\% & 3\% \\
    Deflected answers & 5\% & 15\% \\
    No-preference answers & 15\% & 17\% \\
    \bottomrule
  \end{tabular}
\end{table}

The simulator is instructed to answer only what was asked, one detail at a time, to deflect open-ended questions, and to state no preference where the description is silent. To check that the trained assistants earn their recovery through their questions rather than by exploiting the simulator (e.g., by asking broad questions that draw out several details at once), we audited its answers with a judge (Gemini 3.5 Flash-Lite).

Of the 15 evaluation conditions described in \Cref{sec:analysis}, we audit the two in which the patient-trained and the impatient-trained assistant run in fixed-asking mode under the patient user, the setting in which every question is answered and recovery is highest. This gives one answer per hidden detail of the 78 test instances, i.e., 667 answers per assistant. For each answer, the judge receives the full description and the list of details, and reports which hidden details it states or entails, and whether the question asked for each of them specifically, by topic (i.e., naming the entity or aspect without naming the property), or not at all.

\Cref{tab:audit} summarizes the verdicts. Answers stay on topic: nearly every hidden detail that an answer states was asked for, either specifically or by topic, and volunteered details are rare. This suggests that the trained assistants do not simply exploit the simulator, and that what they recover, they largely recover by asking for it.

\FloatBarrier
\section{User Study Materials}
\label{app:userstudy}

\begin{figure}[!htb]
  \centering
  \includegraphics[width=0.9\linewidth,trim={4cm 0 4cm 3.5cm},clip]{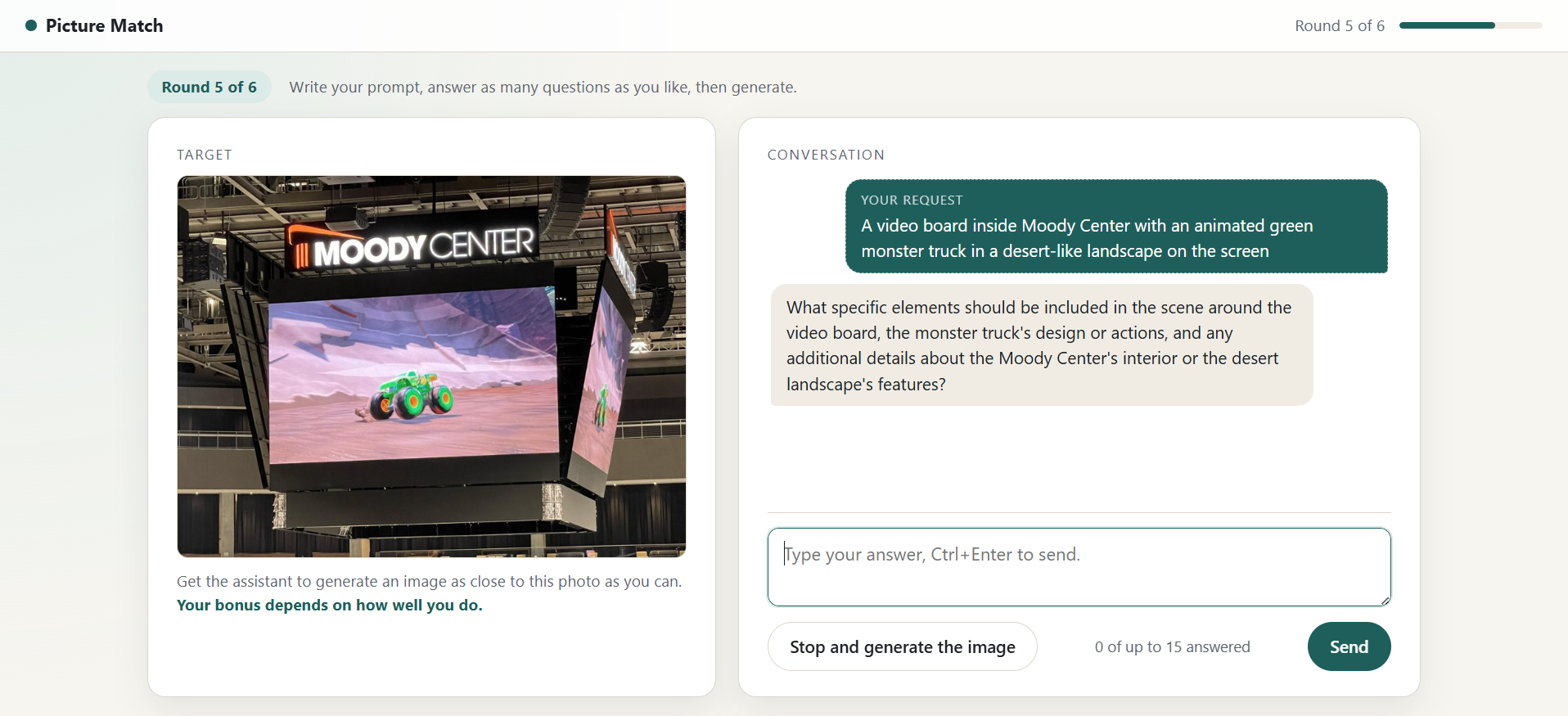}
  \caption{The round screen of the study, with the target on the left and the conversation on the right.}
  \Description{Screenshot of the Picture Match round screen. A target photograph fills the left half under an instruction to get the assistant to generate an image as close to it as possible. The right half is a chat with the participant's request, one question from the assistant, an answer box, a send button and a button to stop and generate the image.}
  \label{fig:interface}
\end{figure}

The study ran as a web application, Picture Match, which participants opened from Prolific with their Prolific ID attached. Below we reproduce what participants read, in the order they read it. \Cref{fig:interface} shows the round screen.

\paragraph{Consent.} The landing page is headed ``Recreate a photo with an AI image tool'' and says: ``You will try to recreate photos with an AI image generator that can ask you questions before it draws. The whole study takes about 45 minutes.'' and, below it, ``The session ends 90 minutes after you start, so please finish it in one sitting.'' The consent form then lists the following:
\begin{itemize}
  \item \textbf{Study purpose.} We study how well AI assistants help people get the image they have in mind.
  \item \textbf{What you will do.} In each round you see a target photo, write a prompt for it the way you would in an image tool, answer the assistant's questions for as long as you like, and receive a generated image.
  \item \textbf{Data collected.} The text you type, the assistant's questions, the generated images, timestamps, your answers to a few screening questions, and your Prolific ID for payment. No other personal information.
  \item \textbf{Duration.} About 45 minutes.
  \item \textbf{Payment.} You receive the base payment for completing the study, plus a bonus that depends on how closely each image you generate matches its target, scored after the study.
  \item \textbf{Voluntary participation.} You may stop at any time by closing this tab, but only completed sessions are compensated.
\end{itemize}
It closes with ``By clicking `I consent' you confirm that you are at least 18 years old, that you have read the information above, and that you agree to take part and to the use of your anonymized responses for research.''

\paragraph{Screening.} Three yes/no questions follow, introduced as ``These decide whether the study will work well for you.'' Answering no to any of them ends the session.
\begin{enumerate}
  \item Are you 18 years old or older?
  \item Are you a native English speaker?
  \item Is your vision normal or corrected to normal (glasses or contacts are fine), including normal color vision?
\end{enumerate}

\paragraph{Instructions.} The instructions are headed ``How it works'':
\begin{itemize}
  \item \textbf{Look at the target.} Each round shows you a photo. Your goal is to get the AI to generate an image as close to that photo as you can.
  \item \textbf{Write a prompt, then answer questions.} Type what you want, the way you would in an image tool. The assistant then asks you questions about the picture, one at a time. Answer from what you see, as specifically as you like.
  \item \textbf{Stop when you are ready.} Click Stop and generate the image whenever you think the assistant knows enough, or keep answering, up to 15 questions. The image is generated once, at the end, and you cannot go back to it.
\end{itemize}
Below them, a highlighted box says ``Your payment depends on how well you perform the task.'', followed by the payment note of the consent form.

\paragraph{Comprehension check.} Participants then answer three multiple-choice items under ``Check your understanding'', which are repeated until all three are correct. The correct option is shown in italics.
\begin{enumerate}
  \item When is the image generated? After every answer I give / \emph{Once, when I stop or the assistant runs out of questions} / Before the assistant asks anything.
  \item Can you stop answering questions before the assistant runs out of them? \emph{Yes, at any time} / No, I have to answer all of them / Only after the first five.
  \item What does your bonus depend on? How many questions I answer / \emph{How closely each generated image matches its target} / How quickly I finish.
\end{enumerate}

\paragraph{Round screen.} \Cref{fig:interface} shows the round screen. The target is shown with the instruction ``Get the assistant to generate an image as close to this photo as you can. Your bonus depends on how well you do.'' The participant writes the request, answers the assistant's questions one at a time, and can stop and generate the image after any question. When the cap is reached, the screen says ``The assistant has no more questions.'' After the image is shown, the screen notes that the match score is calculated after the study. The end screen shows the six target and result pairs, asks ``Anything you want to tell us? (optional)'', and returns the participant to Prolific.

\paragraph{Assistant and image model.} Both models use the ask and compose prompts of \Cref{app:prompts} with the call structure of \Cref{app:training}, with reasoning mode on and 4,096 sampled tokens per call. The decide prompt is not used, since the participant decides when to stop. The final description is rendered with the image model and settings of \Cref{app:training}. Every model call logs its reasoning tokens, its visible tokens, and its latency.

\FloatBarrier
\section{Auto-Interpretability Results}
\label{app:autointerp}

Below we report themes from the Matryoshka summarization~\cite{cherep2026visualpersuasion,sharif2026improving} of the human-study rounds (see \Cref{sec:human_results}). For each reference image played under both models, a language model describes how the trained-model questions, and separately the participants' answers, differ from those of the base model, and these descriptions are then clustered and summarized recursively into top/root-level themes using Gemini 3.8 Flash. \Cref{tab:survey_question_themes} lists themes for the assistant's questions and \Cref{tab:survey_answer_themes} those for the participants' answers, each with example quotations from the trained-model rounds followed by the base-model quotations they are contrasted with. Note: we report second-from-top-level themes, in order to balance capturing variation with meaningful abstraction.

\begin{table*}[!htb]
    \footnotesize
    \caption{Second-from-top-level themes of how the trained vs. base assistant's questions differ.}
    \label{tab:survey_question_themes}
    \begin{tabular}{@{}p{\dimexpr0.23\linewidth-\tabcolsep\relax}p{\dimexpr0.77\linewidth-\tabcolsep\relax}@{}}
\hline
\textbf{Theme} & \textbf{Description}\tabularnewline[0.2em]
\hline
\noalign{\vskip 0.3em}
\raggedright\textbf{More questions asking what should be included} & \raggedright ``What details should be included in the background elements''; ``What specific elements should be included in the scene composition,''; ``What should be included in the depiction of the elephant sculpture'', compared with ``What color are the lanterns?''; ``What color palette or textures should emphasize the urban atmosphere,''; ``What time of day is it during daylight?''.\tabularnewline[0.1em]
\raggedright\textbf{Fewer polar or binary questions} & \raggedright ``What should be included in the background or any other elements''; ``What should be included in the scene besides''; ``What should be included in the foreground'', compared with ``Is the translucent sunlight ray hitting the framed calligraphy''; ``Is the toy car and horse positioned on''; ``Are the castles and purple lions part of the flag's design''.\tabularnewline[0.1em]
\raggedright\textbf{More questions opening with What specific} & \raggedright ``What specific details should be included in the visual style''; ``What specific details should be included for the buildings''; ``What specific elements should be included'', compared with ``What atmosphere or time of day would you like''; ``What time of year is this scene set?''; ``Could you clarify which part of the bed''.\tabularnewline[0.1em]
\raggedright\textbf{Less questions starting with Is or Are} & \raggedright ``What elements should be included in the scene''; ``What should be included in the scene''; ``What should be included in the composition,'', compared with ``Is the sun positioned directly above the main boat''; ``Is the statue meant to represent a Roman or Greek figure''; ``Is the phrase "u can" intended to be "you can"''.\tabularnewline[0.1em]
\hline
\end{tabular}

\end{table*}

\begin{table*}[!htb]
    \footnotesize
    \caption{Second-from-top-level themes of how participants' answers to the trained vs. base assistant differ.}
    \label{tab:survey_answer_themes}
    \begin{tabular}{@{}p{\dimexpr0.23\linewidth-\tabcolsep\relax}p{\dimexpr0.77\linewidth-\tabcolsep\relax}@{}}
\hline
\textbf{Theme} & \textbf{Description}\tabularnewline[0.2em]
\hline
\noalign{\vskip 0.3em}
\raggedright\textbf{Longer full sentences and descriptions} & \raggedright ``Both cats are sitting upright looking forward toward the camera.''; ``A high-resolution, clear, and detailed digital photograph emphasizing natural textures, vibrant colors,''; ``The boat is a fishing boat. There is only rippling water in'', compared with ``Brown table''; ``about the same shade of bright lite green''; ``Looks like a red container''.\tabularnewline[0.1em]
\raggedright\textbf{Less hedging and speculative wording} & \raggedright ``The background is medium colored wood''; ``The foreground includes a paved sidewalk and a dark metal fence''; ``two dark brick chimneys rising from the roof'', compared with ``Im not sure but maybe the left''; ``I think maybe gothic revival''; ``It looks like there could be a chimney''.\tabularnewline[0.1em]
\raggedright\textbf{More spelling mistakes and typos} & \raggedright ``bright perhaps flourescent.''; ``liver shiny lok handle attached to the fence''; ``Th walls are a deep'', compared with ``I purple dial lock behind the pink main one.''; ``stationary decorative sculpture.''; ``standalone carousel rooster sculpture displayed on its own,''.\tabularnewline[0.1em]
\raggedright\textbf{Longer descriptions using complete sentences} & \raggedright ``The composition is almost entirely filled with dense, layered tree branches''; ``The left side of the house has a projecting bay section''; ``The mattress itself is covered with white soft bed sheets and comforter,'', compared with ``warm lights, orange/yellow''; ``light blue''; ``Bed head rest and the front bed edge''.\tabularnewline[0.1em]
\hline
\end{tabular}

\end{table*}

\FloatBarrier
\section{Regression Tables}
\label{app:regressions}

Below we report the full regression tables behind the contrasts of \Cref{sec:results}. All models follow the estimation approach of \Cref{sec:analysis}, and coefficient $p$-values are unadjusted, unlike the BH-adjusted contrasts in the figures.

\subsection{Model Results}
\label{sec:model_results}

\Cref{tab:recovery_regressions,tab:semantic_regressions,tab:crossmodal_regressions,tab:burden_regressions} are the tables for the simulated evaluation, covering recovered utility, the within-modality and cross-modal similarity gains, and the dialogue measures. All use source-image fixed effects and source-image-clustered standard errors, where the base model under fixed asking with the patient user is the reference level. The additive, two-way, and three-way columns add interactions between assistant, asking mode, and user, and the remaining columns add the dialogue measures as covariates.

\begin{table*}[!htb]
    \small
    \caption{Regressions of recovered utility in the simulated evaluation.}
    \label{tab:recovery_regressions}
    {\setlength{\tabcolsep}{2.5pt}
\begin{tabular}{lcccccc}
   \tabularnewline \midrule \midrule
   Dependent Variable: & \multicolumn{6}{c}{Equal-weighted recovery}\\
   Specification                                        & Additive        & Two-way         & Three-way       & + Count        & + Q. length    & + A./output \\   
   Model:                                               & (1)             & (2)             & (3)             & (4)            & (5)            & (6)\\  
   \midrule
   \emph{Variables}\\
   Patient-trained                                      & 0.496$^{****}$  & 0.502$^{****}$  & 0.506$^{****}$  & 0.506$^{****}$ & 0.391$^{****}$ & 0.277$^{****}$\\   
                                                        & (0.024)         & (0.033)         & (0.034)         & (0.034)        & (0.048)        & (0.041)\\   
   Impatient-trained                                    & 0.455$^{****}$  & 0.475$^{****}$  & 0.487$^{****}$  & 0.487$^{****}$ & 0.318$^{****}$ & 0.219$^{****}$\\   
                                                        & (0.022)         & (0.030)         & (0.032)         & (0.032)        & (0.052)        & (0.051)\\   
   Open                                                 & -0.113$^{****}$ & -0.165$^{****}$ & -0.155$^{****}$ & 0.210$^{***}$  & 0.143$^{*}$    & 0.107$^{*}$\\   
                                                        & (0.017)         & (0.024)         & (0.024)         & (0.057)        & (0.055)        & (0.052)\\   
   Impatient                                            & -0.053$^{***}$  & -0.004          & 0.007           & 0.105$^{**}$   & 0.090$^{**}$   & 0.082$^{**}$\\   
                                                        & (0.013)         & (0.024)         & (0.030)         & (0.032)        & (0.033)        & (0.030)\\   
   Patient-trained $\times$ Open                        &                 & 0.078$^{*}$     & 0.072           & -0.210$^{**}$  & -0.146$^{*}$   & -0.105\\   
                                                        &                 & (0.035)         & (0.044)         & (0.063)        & (0.062)        & (0.055)\\   
   Impatient-trained $\times$ Open                      &                 & 0.048           & 0.022           & -0.183$^{**}$  & -0.124$^{*}$   & -0.117$^{*}$\\   
                                                        &                 & (0.033)         & (0.041)         & (0.054)        & (0.057)        & (0.054)\\   
   Patient-trained $\times$ Impatient                   &                 & -0.092$^{**}$   & -0.098$^{*}$    & -0.098$^{*}$   & -0.084         & -0.082$^{*}$\\   
                                                        &                 & (0.034)         & (0.041)         & (0.041)        & (0.043)        & (0.040)\\   
   Impatient-trained $\times$ Impatient                 &                 & -0.086$^{**}$   & -0.112$^{**}$   & -0.112$^{**}$  & -0.105$^{**}$  & -0.106$^{**}$\\   
                                                        &                 & (0.027)         & (0.039)         & (0.040)        & (0.039)        & (0.036)\\   
   Open $\times$ Impatient                              &                 & 0.020           & -0.002          & -0.102$^{**}$  & -0.089$^{*}$   & -0.081$^{*}$\\   
                                                        &                 & (0.029)         & (0.035)         & (0.035)        & (0.036)        & (0.035)\\   
   Patient-trained $\times$ Open $\times$ Impatient     &                 &                 & 0.013           & 0.063          & 0.055          & 0.046\\   
                                                        &                 &                 & (0.058)         & (0.051)        & (0.053)        & (0.050)\\   
   Impatient-trained $\times$ Open $\times$ Impatient   &                 &                 & 0.052           & 0.068          & 0.066          & 0.064\\   
                                                        &                 &                 & (0.055)         & (0.053)        & (0.052)        & (0.048)\\   
   Any answer                                           &                 &                 &                 & 0.121$^{**}$   & 0.017          & -0.117$^{**}$\\   
                                                        &                 &                 &                 & (0.036)        & (0.043)        & (0.041)\\   
   Questions                                            &                 &                 &                 & 0.034$^{****}$ & 0.027$^{****}$ & 0.029$^{****}$\\   
                                                        &                 &                 &                 & (0.004)        & (0.003)        & (0.003)\\   
   Words/question                                       &                 &                 &                 &                & 0.005$^{****}$ & 0.004$^{***}$\\   
                                                        &                 &                 &                 &                & (0.0010)       & (0.0010)\\   
   Words/answer                                         &                 &                 &                 &                &                & 0.013$^{****}$\\   
                                                        &                 &                 &                 &                &                & (0.003)\\   
   Output words / 100                                   &                 &                 &                 &                &                & 0.029\\   
                                                        &                 &                 &                 &                &                & (0.025)\\   
   \midrule
   \emph{Fixed-effects}\\
   Source image                                         & Yes             & Yes             & Yes             & Yes            & Yes            & Yes\\  
   \midrule
   \emph{Fit statistics}\\
   Observations                                         & 936             & 936             & 936             & 936            & 936            & 936\\  
   Within R$^2$                                         & 0.563           & 0.570           & 0.570           & 0.653          & 0.665          & 0.689\\  
   \midrule \midrule
   \multicolumn{7}{l}{\emph{Clustered (Source image) standard-errors in parentheses}}\\
   \multicolumn{7}{l}{\emph{Signif. Codes: ****: 1e-04, ***: 0.001, **: 0.01, *: 0.05}}\\
\end{tabular}
}

\end{table*}

\begin{table*}[!htb]
    \small
    \caption{Regressions of the text-to-text and image-to-image similarity gains in the simulated evaluation.}
    \label{tab:semantic_regressions}
    {\setlength{\tabcolsep}{2.5pt}
\begin{tabular}{lcccccc}
   \tabularnewline \midrule \midrule
   Dependent Variables: & \multicolumn{3}{c}{Text--text} & \multicolumn{3}{c}{Image--image}\\
   Specification                                        & Additive        & Three-way       & + A./output    & Additive        & Three-way      & + A./output \\   
   Model:                                               & (1)             & (2)             & (3)            & (4)             & (5)            & (6)\\  
   \midrule
   \emph{Variables}\\
   Patient-trained                                      & 0.171$^{****}$  & 0.157$^{****}$  & 0.120$^{****}$ & 0.053$^{****}$  & 0.050$^{****}$ & 0.032$^{***}$\\   
                                                        & (0.014)         & (0.015)         & (0.016)        & (0.006)         & (0.007)        & (0.008)\\   
   Impatient-trained                                    & 0.100$^{****}$  & 0.112$^{****}$  & 0.090$^{****}$ & 0.029$^{****}$  & 0.032$^{****}$ & 0.017\\   
                                                        & (0.012)         & (0.015)         & (0.018)        & (0.005)         & (0.007)        & (0.009)\\   
   Open                                                 & -0.042$^{****}$ & -0.058$^{****}$ & 0.019          & -0.016$^{****}$ & -0.021$^{***}$ & 0.006\\   
                                                        & (0.006)         & (0.011)         & (0.018)        & (0.004)         & (0.006)        & (0.009)\\   
   Impatient                                            & -0.012$^{**}$   & 0.006           & 0.039$^{****}$ & -0.005$^{*}$    & 0.004          & 0.018$^{***}$\\   
                                                        & (0.004)         & (0.009)         & (0.009)        & (0.002)         & (0.004)        & (0.005)\\   
   Patient-trained $\times$ Open                        &                 & 0.038$^{**}$    & -0.018         &                 & 0.013          & -0.006\\   
                                                        &                 & (0.013)         & (0.018)        &                 & (0.007)        & (0.009)\\   
   Impatient-trained $\times$ Open                      &                 & 0.002           & -0.022         &                 & 0.002          & -0.003\\   
                                                        &                 & (0.013)         & (0.021)        &                 & (0.008)        & (0.011)\\   
   Patient-trained $\times$ Impatient                   &                 & -0.018          & -0.017         &                 & -0.011         & -0.010\\   
                                                        &                 & (0.009)         & (0.009)        &                 & (0.006)        & (0.006)\\   
   Impatient-trained $\times$ Impatient                 &                 & -0.045$^{***}$  & -0.044$^{***}$ &                 & -0.017$^{**}$  & -0.017$^{**}$\\   
                                                        &                 & (0.012)         & (0.012)        &                 & (0.006)        & (0.006)\\   
   Open $\times$ Impatient                              &                 & -0.012          & -0.046$^{***}$ &                 & -0.006         & -0.021$^{***}$\\   
                                                        &                 & (0.010)         & (0.011)        &                 & (0.005)        & (0.006)\\   
   Patient-trained $\times$ Open $\times$ Impatient     &                 & 0.017           & 0.033$^{*}$    &                 & 0.007          & 0.014\\   
                                                        &                 & (0.015)         & (0.014)        &                 & (0.008)        & (0.009)\\   
   Impatient-trained $\times$ Open $\times$ Impatient   &                 & 0.038$^{*}$     & 0.042$^{*}$    &                 & 0.017          & 0.019$^{*}$\\   
                                                        &                 & (0.017)         & (0.016)        &                 & (0.010)        & (0.009)\\   
   Any answer                                           &                 &                 & -0.060$^{*}$   &                 &                & -0.035$^{**}$\\   
                                                        &                 &                 & (0.024)        &                 &                & (0.011)\\   
   Questions                                            &                 &                 & 0.013$^{****}$ &                 &                & 0.005$^{****}$\\   
                                                        &                 &                 & (0.001)        &                 &                & (0.0008)\\   
   Words/question                                       &                 &                 & 0.0003         &                 &                & 0.0002\\   
                                                        &                 &                 & (0.0004)       &                 &                & (0.0002)\\   
   Words/answer                                         &                 &                 & 0.003$^{**}$   &                 &                & 0.001$^{**}$\\   
                                                        &                 &                 & (0.0010)       &                 &                & (0.0004)\\   
   Output words / 100                                   &                 &                 & -0.008         &                 &                & -0.0004\\   
                                                        &                 &                 & (0.011)        &                 &                & (0.005)\\   
   \midrule
   \emph{Fixed-effects}\\
   Source image                                         & Yes             & Yes             & Yes            & Yes             & Yes            & Yes\\  
   \midrule
   \emph{Fit statistics}\\
   Observations                                         & 936             & 936             & 936            & 936             & 936            & 936\\  
   Within R$^2$                                         & 0.476           & 0.487           & 0.578          & 0.272           & 0.282          & 0.386\\  
   \midrule \midrule
   \multicolumn{7}{l}{\emph{Clustered (Source image) standard-errors in parentheses}}\\
   \multicolumn{7}{l}{\emph{Signif. Codes: ****: 1e-04, ***: 0.001, **: 0.01, *: 0.05}}\\
\end{tabular}
}

\end{table*}

\begin{table*}[!htb]
    \small
    \caption{Regressions of the text-to-image and image-to-text similarity gains in the simulated evaluation.}
    \label{tab:crossmodal_regressions}
    {\setlength{\tabcolsep}{2.5pt}
\begin{tabular}{lcccccc}
   \tabularnewline \midrule \midrule
   Dependent Variables: & \multicolumn{3}{c}{Text--image} & \multicolumn{3}{c}{Image--text}\\
   Specification                                        & Additive        & Three-way      & + A./output           & Additive        & Three-way      & + A./output \\   
   Model:                                               & (1)             & (2)            & (3)                   & (4)             & (5)            & (6)\\  
   \midrule
   \emph{Variables}\\
   Patient-trained                                      & 0.066$^{****}$  & 0.057$^{****}$ & 0.043$^{****}$        & 0.079$^{****}$  & 0.074$^{****}$ & 0.046$^{****}$\\   
                                                        & (0.007)         & (0.007)        & (0.008)               & (0.008)         & (0.008)        & (0.009)\\   
   Impatient-trained                                    & 0.042$^{****}$  & 0.044$^{****}$ & 0.038$^{****}$        & 0.054$^{****}$  & 0.059$^{****}$ & 0.033$^{**}$\\   
                                                        & (0.006)         & (0.007)        & (0.009)               & (0.007)         & (0.008)        & (0.010)\\   
   Open                                                 & -0.017$^{****}$ & -0.025$^{***}$ & -0.0009               & -0.021$^{****}$ & -0.027$^{***}$ & 0.001\\   
                                                        & (0.003)         & (0.006)        & (0.010)               & (0.004)         & (0.007)        & (0.011)\\   
   Impatient                                            & -0.004$^{*}$    & 0.002          & 0.017$^{***}$         & -0.007$^{**}$   & 0.004          & 0.021$^{***}$\\   
                                                        & (0.002)         & (0.004)        & (0.004)               & (0.003)         & (0.005)        & (0.005)\\   
   Patient-trained $\times$ Open                        &                 & 0.019$^{**}$   & 0.002                 &                 & 0.017$^{*}$    & -0.0010\\   
                                                        &                 & (0.007)        & (0.010)               &                 & (0.009)        & (0.011)\\   
   Impatient-trained $\times$ Open                      &                 & 0.004          & 0.003                 &                 & 0.0005         & -0.0007\\   
                                                        &                 & (0.007)        & (0.012)               &                 & (0.010)        & (0.014)\\   
   Patient-trained $\times$ Impatient                   &                 & -0.004         & -0.003                &                 & -0.013         & -0.011\\   
                                                        &                 & (0.005)        & (0.005)               &                 & (0.007)        & (0.007)\\   
   Impatient-trained $\times$ Impatient                 &                 & -0.017$^{***}$ & -0.017$^{**}$         &                 & -0.020$^{**}$  & -0.020$^{**}$\\   
                                                        &                 & (0.005)        & (0.005)               &                 & (0.007)        & (0.006)\\   
   Open $\times$ Impatient                              &                 & -0.007         & -0.021$^{***}$        &                 & -0.011         & -0.027$^{***}$\\   
                                                        &                 & (0.005)        & (0.006)               &                 & (0.006)        & (0.007)\\   
   Patient-trained $\times$ Open $\times$ Impatient     &                 & 0.004          & 0.012                 &                 & 0.011          & 0.018\\   
                                                        &                 & (0.007)        & (0.007)               &                 & (0.010)        & (0.010)\\   
   Impatient-trained $\times$ Open $\times$ Impatient   &                 & 0.019$^{*}$    & 0.021$^{**}$          &                 & 0.020          & 0.022$^{*}$\\   
                                                        &                 & (0.008)        & (0.007)               &                 & (0.012)        & (0.011)\\   
   Any answer                                           &                 &                & -0.036$^{**}$         &                 &                & -0.050$^{***}$\\   
                                                        &                 &                & (0.012)               &                 &                & (0.014)\\   
   Questions                                            &                 &                & 0.005$^{****}$        &                 &                & 0.006$^{****}$\\   
                                                        &                 &                & (0.0008)              &                 &                & (0.0009)\\   
   Words/question                                       &                 &                & $8.09\times 10^{-5}$  &                 &                & 0.0004\\   
                                                        &                 &                & (0.0002)              &                 &                & (0.0002)\\   
   Words/answer                                         &                 &                & 0.001$^{**}$          &                 &                & 0.002$^{***}$\\   
                                                        &                 &                & (0.0004)              &                 &                & (0.0005)\\   
   Output words / 100                                   &                 &                & -0.004                &                 &                & -0.0008\\   
                                                        &                 &                & (0.005)               &                 &                & (0.006)\\   
   \midrule
   \emph{Fixed-effects}\\
   Source image                                         & Yes             & Yes            & Yes                   & Yes             & Yes            & Yes\\  
   \midrule
   \emph{Fit statistics}\\
   Observations                                         & 936             & 936            & 936                   & 936             & 936            & 936\\  
   Within R$^2$                                         & 0.356           & 0.368          & 0.451                 & 0.367           & 0.376          & 0.469\\  
   \midrule \midrule
   \multicolumn{7}{l}{\emph{Clustered (Source image) standard-errors in parentheses}}\\
   \multicolumn{7}{l}{\emph{Signif. Codes: ****: 1e-04, ***: 0.001, **: 0.01, *: 0.05}}\\
\end{tabular}
}

\end{table*}

\begin{table*}[!htb]
    \small
    \caption{Regressions of the dialogue measures in the simulated evaluation.}
    \label{tab:burden_regressions}
    {\setlength{\tabcolsep}{2.5pt}
\begin{tabular}{lcccccc}
   \tabularnewline \midrule \midrule
   Dependent Variables: & Questions & \multicolumn{2}{c}{Question words} & \multicolumn{2}{c}{Answer words} & Output words\\
   Specification & \multicolumn{2}{c}{Additive} & Three-way & Additive & \multicolumn{2}{c}{Three-way} \\ 
   Model:                                               & (1)            & (2)             & (3)             & (4)            & (5)            & (6)\\  
   \midrule
   \emph{Variables}\\
   Patient-trained                                      & 2.81$^{****}$  & 218.9$^{****}$  & 219.3$^{****}$  & 82.1$^{****}$  & 82.4$^{****}$  & 4.05\\   
                                                        & (0.201)        & (14.8)          & (16.4)          & (4.62)         & (5.77)         & (3.30)\\   
   Impatient-trained                                    & 1.63$^{****}$  & 218.0$^{****}$  & 314.4$^{****}$  & 50.1$^{****}$  & 54.0$^{****}$  & 130.7$^{****}$\\   
                                                        & (0.076)        & (10.0)          & (20.7)          & (2.05)         & (4.02)         & (4.92)\\   
   Open                                                 & -3.82$^{****}$ & -126.6$^{****}$ & -155.5$^{****}$ & -32.0$^{****}$ & -51.2$^{****}$ & -2.27\\   
                                                        & (0.326)        & (15.0)          & (12.5)          & (3.72)         & (3.56)         & (2.99)\\   
   Impatient                                            & -1.78$^{****}$ & -80.0$^{****}$  & -61.6$^{****}$  & -19.0$^{****}$ & -12.4$^{***}$  & -2.51\\   
                                                        & (0.213)        & (10.0)          & (11.8)          & (2.87)         & (3.15)         & (2.90)\\   
   Patient-trained $\times$ Open                        &                &                 & 83.1$^{*}$      &                & 27.2$^{*}$     & 0.410\\   
                                                        &                &                 & (33.5)          &                & (10.7)         & (4.64)\\   
   Impatient-trained $\times$ Open                      &                &                 & -127.6$^{****}$ &                & 3.21           & -0.385\\   
                                                        &                &                 & (26.6)          &                & (5.63)         & (6.89)\\   
   Patient-trained $\times$ Impatient                   &                &                 & -80.5$^{****}$  &                & -31.5$^{****}$ & -2.94\\   
                                                        &                &                 & (15.1)          &                & (5.79)         & (4.18)\\   
   Impatient-trained $\times$ Impatient                 &                &                 & -106.0$^{****}$ &                & -15.9$^{***}$  & 0.923\\   
                                                        &                &                 & (19.7)          &                & (3.94)         & (7.39)\\   
   Open $\times$ Impatient                              &                &                 & 62.6$^{****}$   &                & 12.6$^{***}$   & 4.13\\   
                                                        &                &                 & (11.8)          &                & (3.23)         & (3.62)\\   
   Patient-trained $\times$ Open $\times$ Impatient     &                &                 & -6.65           &                & 7.27           & -5.96\\   
                                                        &                &                 & (42.8)          &                & (12.6)         & (5.70)\\   
   Impatient-trained $\times$ Open $\times$ Impatient   &                &                 & 81.6$^{**}$     &                & 9.74           & -10.1\\   
                                                        &                &                 & (28.2)          &                & (6.02)         & (9.45)\\   
   \midrule
   \emph{Fixed-effects}\\
   Source image                                         & Yes            & Yes             & Yes             & Yes            & Yes            & Yes\\  
   \midrule
   \emph{Fit statistics}\\
   Observations                                         & 936            & 936             & 936             & 936            & 936            & 936\\  
   Within R$^2$                                         & 0.431          & 0.445           & 0.502           & 0.503          & 0.536          & 0.824\\  
   \midrule \midrule
   \multicolumn{7}{l}{\emph{Clustered (Source image) standard-errors in parentheses}}\\
   \multicolumn{7}{l}{\emph{Signif. Codes: ****: 1e-04, ***: 0.001, **: 0.01, *: 0.05}}\\
\end{tabular}
}

\end{table*}

\FloatBarrier
\subsection{Human Results}

\Cref{tab:quality_text_cosine_gain_regressions,tab:quality_image_cosine_gain_regressions,tab:quality_text_image_cosine_gain_regressions,tab:quality_image_text_cosine_gain_regressions} are the tables for the four similarity gains in the human study, without and with adjustment for questions answered and source-image fixed effects, and \Cref{tab:process_regressions} those for its process measures. All use participant-clustered SEs.

\begin{table*}[!htb]
    \small
    \begin{minipage}[t]{0.49\textwidth}
        \centering
        \caption{Human-study regressions ($\Delta$ text--text).}
        \label{tab:quality_text_cosine_gain_regressions}
        \begin{tabular}{lccc}
   \tabularnewline \midrule \midrule
   Outcome & \multicolumn{3}{c}{Text--text} \\ 
   Specification      & Training    & Adjusted      & Adjusted + FE \\   
   Model:             & (1)         & (2)           & (3)\\  
   \midrule
   \emph{Variables}\\
   Constant           & 0.018$^{*}$ & -0.013        &   \\   
                      & (0.008)     & (0.010)       &   \\   
   Trained model      & 0.017$^{*}$ & 0.030$^{***}$ & 0.034$^{***}$\\   
                      & (0.007)     & (0.008)       & (0.009)\\   
   Questions answered &             & 0.006$^{**}$  & 0.007$^{**}$\\   
                      &             & (0.002)       & (0.002)\\   
   \midrule
   \emph{Fixed-effects}\\
   Item               &             &               & Yes\\  
   \midrule
   \emph{Fit statistics}\\
   Observations       & 456         & 456           & 456\\  
   R$^2$              & 0.009       & 0.059         & 0.199\\  
   \midrule \midrule
   \multicolumn{4}{l}{\emph{Clustered (Participant ID) standard-errors in parentheses}}\\
   \multicolumn{4}{l}{\emph{Signif. Codes: ****: 1e-04, ***: 0.001, **: 0.01, *: 0.05}}\\
\end{tabular}

    \end{minipage}
    \hfill
    \begin{minipage}[t]{0.49\textwidth}
        \centering
        \caption{Human-study regressions ($\Delta$ image--image).}
        \label{tab:quality_image_cosine_gain_regressions}
        \begin{tabular}{lccc}
   \tabularnewline \midrule \midrule
   Outcome & \multicolumn{3}{c}{Image--image} \\ 
   Specification      & Training    & Adjusted       & Adjusted + FE \\   
   Model:             & (1)         & (2)            & (3)\\  
   \midrule
   \emph{Variables}\\
   Constant           & 0.008$^{*}$ & -0.008$^{*}$   &   \\   
                      & (0.003)     & (0.004)        &   \\   
   Trained model      & 0.007$^{*}$ & 0.015$^{****}$ & 0.014$^{**}$\\   
                      & (0.003)     & (0.003)        & (0.005)\\   
   Questions answered &             & 0.003$^{****}$ & 0.004$^{****}$\\   
                      &             & (0.0008)       & (0.0009)\\   
   \midrule
   \emph{Fixed-effects}\\
   Item               &             &                & Yes\\  
   \midrule
   \emph{Fit statistics}\\
   Observations       & 456         & 456            & 456\\  
   R$^2$              & 0.008       & 0.070          & 0.238\\  
   \midrule \midrule
   \multicolumn{4}{l}{\emph{Clustered (Participant ID) standard-errors in parentheses}}\\
   \multicolumn{4}{l}{\emph{Signif. Codes: ****: 1e-04, ***: 0.001, **: 0.01, *: 0.05}}\\
\end{tabular}

    \end{minipage}
\end{table*}

\begin{table*}[!htb]
    \small
    \begin{minipage}[t]{0.49\textwidth}
        \centering
        \caption{Human-study regressions, $\Delta$ text--image similarity.}
        \label{tab:quality_text_image_cosine_gain_regressions}
        \begin{tabular}{lccc}
   \tabularnewline \midrule \midrule
   Outcome & \multicolumn{3}{c}{Text--image} \\ 
   Specification      & Training        & Adjusted        & Adjusted + FE \\   
   Model:             & (1)             & (2)             & (3)\\  
   \midrule
   \emph{Variables}\\
   Constant           & -0.020$^{****}$ & -0.030$^{****}$ &   \\   
                      & (0.003)         & (0.004)         &   \\   
   Trained model      & 0.010$^{***}$   & 0.015$^{****}$  & 0.015$^{**}$\\   
                      & (0.003)         & (0.004)         & (0.005)\\   
   Questions answered &                 & 0.002$^{**}$    & 0.003$^{**}$\\   
                      &                 & (0.0007)        & (0.0008)\\   
   \midrule
   \emph{Fixed-effects}\\
   Item               &                 &                 & Yes\\  
   \midrule
   \emph{Fit statistics}\\
   Observations       & 456             & 456             & 456\\  
   R$^2$              & 0.016           & 0.041           & 0.234\\  
   \midrule \midrule
   \multicolumn{4}{l}{\emph{Clustered (Participant ID) standard-errors in parentheses}}\\
   \multicolumn{4}{l}{\emph{Signif. Codes: ****: 1e-04, ***: 0.001, **: 0.01, *: 0.05}}\\
\end{tabular}

    \end{minipage}
    \hfill
    \begin{minipage}[t]{0.49\textwidth}
        \centering
        \caption{Human-study regressions, $\Delta$ image--text similarity.}
        \label{tab:quality_image_text_cosine_gain_regressions}
        \begin{tabular}{lccc}
   \tabularnewline \midrule \midrule
   Outcome & \multicolumn{3}{c}{Image--text} \\ 
   Specification      & Training     & Adjusted       & Adjusted + FE \\   
   Model:             & (1)          & (2)            & (3)\\  
   \midrule
   \emph{Variables}\\
   Constant           & 0.011$^{**}$ & -0.008         &   \\   
                      & (0.004)      & (0.005)        &   \\   
   Trained model      & 0.014$^{**}$ & 0.022$^{****}$ & 0.023$^{***}$\\   
                      & (0.004)      & (0.005)        & (0.006)\\   
   Questions answered &              & 0.004$^{****}$ & 0.004$^{****}$\\   
                      &              & (0.0010)       & (0.001)\\   
   \midrule
   \emph{Fixed-effects}\\
   Item               &              &                & Yes\\  
   \midrule
   \emph{Fit statistics}\\
   Observations       & 456          & 456            & 456\\  
   R$^2$              & 0.018        & 0.075          & 0.234\\  
   \midrule \midrule
   \multicolumn{4}{l}{\emph{Clustered (Participant ID) standard-errors in parentheses}}\\
   \multicolumn{4}{l}{\emph{Signif. Codes: ****: 1e-04, ***: 0.001, **: 0.01, *: 0.05}}\\
\end{tabular}

    \end{minipage}
\end{table*}

\begin{table*}[!htb]
    \small
    \caption{Human-study regressions, process measures.}
    \label{tab:process_regressions}
    {\setlength{\tabcolsep}{2.5pt}
\begin{tabular}{lcccc}
   \tabularnewline \midrule \midrule
   Outcome            & Questions      & Question words & Round seconds & Sampled tokens \\   
   Model:             & (1)            & (2)            & (3)           & (4)\\  
   \midrule
   \emph{Variables}\\
   Trained model      & -2.27$^{****}$ & 111.9$^{****}$ & 2.00          & -1,420.4$^{****}$\\   
                      & (0.277)        & (12.2)         & (30.8)        & (137.9)\\   
   Questions answered &                & 31.3$^{****}$  & 46.2$^{****}$ & 702.5$^{****}$\\   
                      &                & (2.48)         & (6.10)        & (28.6)\\   
   \midrule
   \emph{Fixed-effects}\\
   Source image       & Yes            & Yes            & Yes           & Yes\\  
   \midrule
   \emph{Fit statistics}\\
   Observations       & 456            & 456            & 456           & 456\\  
   Within R$^2$       & 0.117          & 0.733          & 0.261         & 0.827\\  
   \midrule \midrule
   \multicolumn{5}{l}{\emph{Clustered (Participant ID) standard-errors in parentheses}}\\
   \multicolumn{5}{l}{\emph{Signif. Codes: ****: 1e-04, ***: 0.001, **: 0.01, *: 0.05}}\\
\end{tabular}
}

\end{table*}

\FloatBarrier
\section{Additional Results}
\label{app:results}

\begin{figure*}[!htb]
    \centering
    \includegraphics[width=0.9\linewidth]{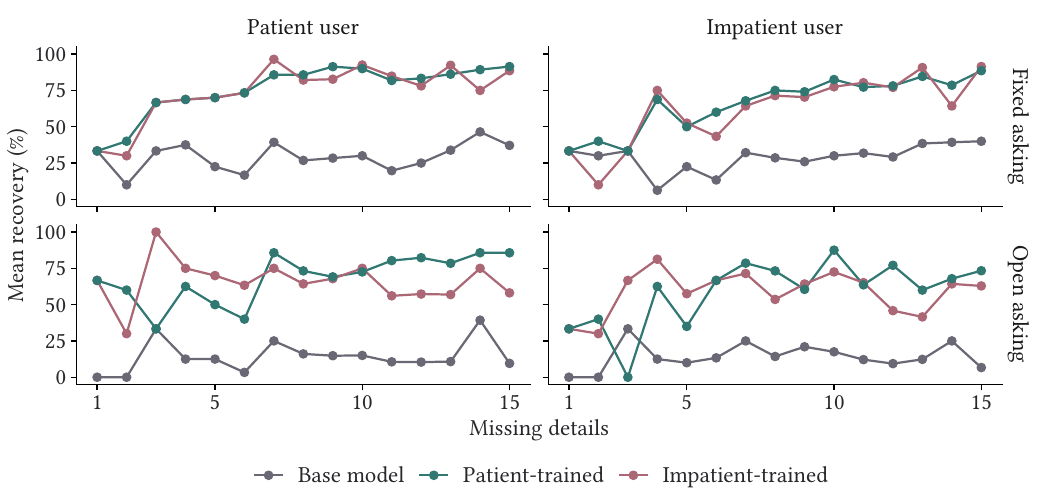}
    \caption{Mean recovery on the test split by the number of hidden details, for each assistant, asking mode and simulated user.}
    \Description{Four panels, fixed and open asking by patient and impatient user, with mean recovery against the number of missing details from 1 to 15. Under fixed asking the two trained lines rise from about 35 percent at one detail to about 90 percent at fifteen, while the base line stays between 10 and 45 percent. Under open asking the trained lines are noisier, between 30 and 100 percent, and the base line stays mostly below 25 percent.}
    \label{fig:task_difficulty}
\end{figure*}

\begin{figure*}[!htb]
    \centering
    \includegraphics[width=\linewidth]{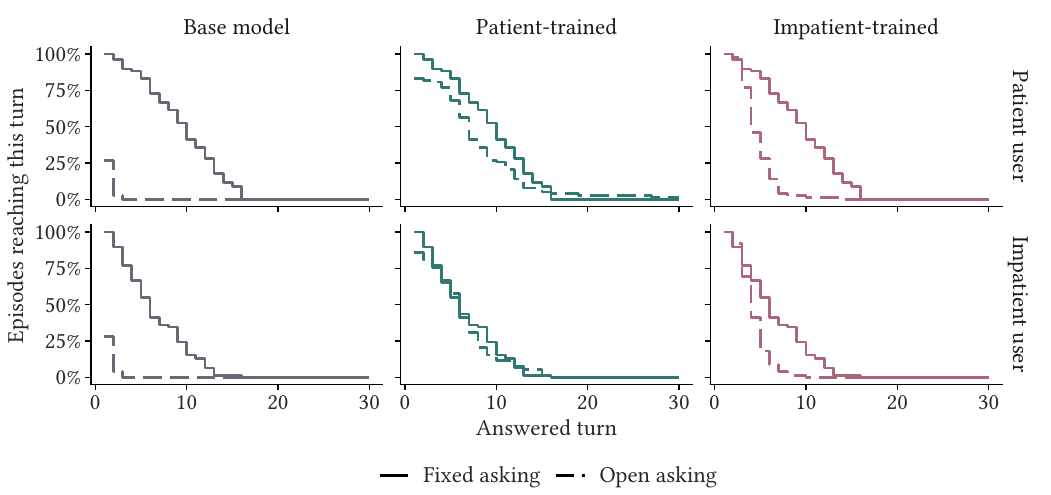}
    \caption{Share of test episodes reaching each answered turn under fixed asking and open asking, for each assistant and simulated user.}
    \Description{Six panels, three assistants by two simulated users, each showing the share of episodes reaching each answered turn as a step curve for fixed and open asking. Fixed-asking curves decline steadily to zero by turn 16. The base model's open-asking curve starts below 30 percent and reaches zero by turn 3. The trained open-asking curves start above 80 percent and fall below the fixed-asking curves, the impatient-trained one earliest.}
    \label{fig:answered_turn_coverage}
\end{figure*}

Here we report the simulated evaluation broken out by the number of hidden details and by dialogue turn, following the evaluation protocol of \Cref{sec:analysis}. \Cref{fig:task_difficulty} disaggregates the recovery of \Cref{fig:recovery} by the number of hidden details $k$, for each assistant, asking mode, and simulated user. \Cref{fig:answered_turn_coverage} shows the distribution of dialogue length as the share of episodes that reach each answered turn, under fixed and open asking. Under open asking, the trained assistants ask in most episodes but stop before the fixed-asking budget, which is the under-asking discussed in \Cref{sec:model_results}.

\FloatBarrier
\section{Prompts}
\label{app:prompts}

Here we show all the prompts for reproducibility.

\subsection{Assistant}

\begin{promptbox}[Ask prompt]
You are an assistant preparing to generate an image for a user.\\
The user's request may be missing details that matter to them.\\
Ask one short clarifying question about whatever would matter most to the user and is still unclear.\\
Don't ask about anything they already told you or anything obvious.\\
Reply with the question only.
\end{promptbox}

\begin{promptbox}[Decide prompt]
You are an assistant preparing to generate an image for a user.\\
The user's request may be missing details that matter to them.\\
Decide whether to ask one more clarifying question before making the image.\\
If anything that could matter to the user is still unclear, answer yes. If not, answer no.\\
Reply with exactly one word: yes or no.
\end{promptbox}

\begin{promptbox}[Compose prompt]
Write the complete, detailed description of the image implied by the user's request and the conversation.\\
Include every detail the user stated, none may be dropped or altered.\\
Where something visually important was never specified, fill in the most likely choice so the description is complete and concrete.\\
Write flowing prose, 100-180 words, like a dense image caption.\\
Reply with the description only.
\end{promptbox}

\subsection{User Simulator}

\begin{promptbox}[User simulator prompt]
You are role-playing a user who asked an assistant to generate an image.\\
The complete ground-truth description of the image you want is given as INTENT.\\
The assistant may ask you clarification questions.

Rules:
\begin{itemize}
  \item Answer only what was asked, truthfully based on the INTENT, in 1-2 short casual sentences.
  \item Never volunteer details that were not asked about.
  \item If the question asks about several specific things, answer only the single most relevant one.
  \item If the question is open-ended and does not ask about anything specific, do not reveal any detail, just say ``Ask me about one specific thing.''
  \item If the INTENT does not specify the answer, say you have no preference.
  \item Never mention the INTENT or that you are role-playing.
\end{itemize}

Expect the following inputs:

\texttt{INTENT}: The complete ground-truth specification of what the user wants.

\texttt{CONVERSATION}: The conversation so far, with the user's original request and any question-answer turns.

\texttt{QUESTION}: The assistant's clarification question to answer.
\tcbline
\emph{Output.} \texttt{answer}: the user's answer to the question.
\end{promptbox}

\subsection{Judge}

\begin{promptbox}[Judge prompt]
You verify whether a piece of text satisfies a specific yes/no check about an image description.

Rules:
\begin{itemize}
  \item Answer true only if the text explicitly states the fact or clearly entails it, paraphrases and synonyms count.
  \item If the text is silent, contradicts it, or is only vaguely compatible, answer false.
  \item Judge only against the given text, do not use outside knowledge about what is likely.
\end{itemize}

Expect the following inputs:

\texttt{CHECK}: The yes/no question about the text to verify.

\texttt{TEXT}: The only text to judge the check against.
\tcbline
\emph{Output.} \texttt{satisfied}: whether the text states or clearly entails the fact.
\end{promptbox}

\subsection{Image Model}

\begin{promptbox}[Image model instruction]
Return a single image as output, do not give any options.
\end{promptbox}

\subsection{Synthesis}

\begin{promptbox}[Decomposition prompt]
Given a complete, detailed description of an image, decompose it into a list of atomic details.

Rules:
\begin{itemize}
  \item Each detail states exactly one checkable fact about the image. Split compound sentences.
  \item Each detail must be self-contained and objectively verifiable from the description alone.
  \item The category must be one of the allowed categories below. Prefer the most precise label.
  \item importance is how much the image would change if this detail were different, from 0 (barely noticeable) to 1 (defining feature).
  \item check is a self-contained yes/no question a judge can ask about any text describing this image.
  \item Every concrete visual fact appears in exactly one detail. Do not invent facts.
\end{itemize}

Allowed categories:
\begin{itemize}
  \item entity: an object or subject is present in the scene.
  \item attribute: a visual property (color, material, texture, size, shape, text content) of a named entity.
  \item relation: a spatial or interactional relation between two named entities.
  \item style: a global rendering or photographic property (medium, lighting, palette, camera angle).
  \item count: an explicit quantity of some entity.
  \item layout: global composition or positioning within the frame.
\end{itemize}

Expect the following inputs:

\texttt{TEXT}: The complete, fully-specified text to decompose.
\tcbline
\emph{Output.} \texttt{details}: every concrete fact in the text, one atomic detail each, where each detail has \texttt{text} (the detail as one self-contained factual sentence naming its entities explicitly), \texttt{category} (one of the allowed categories), \texttt{importance} (how much the intended output would change if this detail were different, 0 to 1) and \texttt{check} (a self-contained yes/no question verifying that a text states this detail).
\end{promptbox}

\begin{promptbox}[Redaction prompt]
You are given the complete description of an image and a list of details to remove from it.\\
Remove those details and write what remains as the request that a user who wants this image would plausibly type into an image-generation tool.

Rules:
\begin{itemize}
  \item Remove every listed detail completely: the request must not state, entail, or hint at any of them.
  \item If removing a detail leaves another part of the description depending on it, rephrase that part so it reads naturally without revealing the removed detail.
  \item Keep every fact of the description that is not listed for removal. Do not add, infer, or compensate for what was removed.
  \item Write a natural, flowing request of 1-3 sentences, the way a real user writes, not a list or caption.
  \item Keep the register casual and compact. Drop trivially inferable connective phrasing, but never a whole fact.
  \item Output only the request text.
\end{itemize}

Expect the following inputs:

\texttt{DESCRIPTION}: The complete, fully-specified original description.

\texttt{REDACTED}: The details to remove, none of them may survive into the request.
\tcbline
\emph{Output.} \texttt{request}: natural 1-3 sentence user request stating everything in the description except the redacted details.
\end{promptbox}

\end{document}